%% file: main.tex
\documentclass[journal]{IEEEtran}

\usepackage{cite}
\usepackage[pdftex]{graphicx}
\graphicspath{{./figures/}}
\usepackage{amsmath}
\usepackage{listings}
\usepackage{xcolor}
\usepackage{url}
\usepackage{enumitem}
\usepackage{subfigure}
\usepackage{stfloats}
\usepackage{caption}
\usepackage{dsfont}
\usepackage{array}
\usepackage{multirow}
\usepackage{booktabs}
\usepackage{hyperref}

\input{definitions}

\begin{document}
\title{
Evaluating Self and Semi-Supervised Methods for Remote Sensing Segmentation Tasks
}



\author{Chaitanya~Patel$^*$,
        Shashank~Sharma$^*$,
        Valerie~J.~Pasquarella,
        Varun~Gulshan
\thanks{$^*$C. Patel and S. Sharma contributed equally as first authors. This work was done when they were with Google Research India. (email: \href{mailto:chaitanya100100@gmail.com}{chaitanya100100@gmail.com}, \href{mailto:shashank.879@gmail.com}{shashank.879@gmail.com})}
\thanks{V. Gulshan is with Google Research India. (email: \href{mailto:varungulshan@google.com}{varungulshan@google.com})}%
\thanks{V.J. Pasquarella is with Google LLC. (email: \href{mailto:valpasq@google.com}{valpasq@google.com})}%
}


\maketitle
\begin{abstract}
Self- and semi-supervised machine learning techniques leverage unlabeled data for improving downstream task performance. These methods are especially valuable for remote sensing tasks where producing labeled ground truth datasets can be prohibitively expensive but there is easy access to a wealth of unlabeled imagery. We perform a rigorous evaluation of SimCLR, a self-supervised method, and FixMatch, a  semi-supervised method, on three remote sensing tasks: riverbed segmentation, land cover mapping, and flood mapping. We quantify performance improvements on these remote sensing segmentation tasks when additional imagery outside of the original supervised dataset is made available for training. We also design experiments to test the effectiveness of these techniques when the test set is domain shifted to sample different geographic areas compared to the training and validation sets. We find that such techniques significantly improve generalization performance when labeled data is limited and there are geographic domain shifts between the training data and the validation/test data.
\end{abstract}


\IEEEpeerreviewmaketitle

\section{Introduction}
\label{sec:introduction}
\input{sections/introduction}

\section{Methods}
\label{sec:methods}
\input{sections/methods}

\section{Datasets}
\label{sec:datasets}
\input{sections/datasets}

\section{Experiments and Results}
\label{sec:experiments_results}
\input{sections/experiments_results}

\section{Conclusion}
\label{sec:conclusion}
\input{sections/conclusion}

\ifCLASSOPTIONcaptionsoff
  \newpage
\fi

\bibliographystyle{IEEEtran}
\bibliography{main}

\newpage

\input{sections/appendix}

\end{document}

%% file: definitions.tex
\newcommand{\supimgnet}[0]{ \textsl{Supervised(ImageNet)} }
\newcommand{\suprandom}[0]{ \textsl{Supervised(Random)} }
\newcommand{\supsimclr}[0]{ \textsl{Supervised(SimCLR)} }

\newcommand{\fmimgnet}[0]{ \textsl{FixMatch(ImageNet)} }
\newcommand{\fmrandom}[0]{ \textsl{FixMatch(Random)} }
\newcommand{\fmsimclr}[0]{ \textsl{FixMatch(SimCLR)} }

\newcommand{\ra}[1]{\renewcommand{\arraystretch}{#1}}
\newcommand{\specialcell}[2][c]{\begin{tabular}[#1]{@{}l@{}}#2\end{tabular}}
\newcommand{\specialcellmid}[2][c]{\begin{tabular}[#1]{@{}c@{}}#2\end{tabular}}

%% file: sections/introduction.tex
\IEEEPARstart{I}{n} recent years, many machine learning (ML) methods that leverage unlabeled data to build high performing models using only small labeled datasets have been proposed \cite{radford2018improving, sohn2020fixmatch, chen2020simple}.
The development of such techniques is important for advancing our understanding of how humans learn (e.g. we do not need millions of labels to learn new concepts) as well as practical applications where only a limited amount of high-quality labeled data is available. Remote sensing applications that use machine learning methods are especially good candidates for leveraging unlabeled data for several reasons: 
(i) labeling can be expensive for certain applications, requiring remote sensing experts (as opposed to a non-expert labeling workforce) or ground surveying (e.g. crop type labeling  \cite{zhong2019}),
(ii) generalization issues/domain shifts are common as the geographical context changes and often require collection of additional labels when deploying models to a new region of interest, and
(iii) there is easy access to freely available imagery (such as the Sentinel-1 and National Agriculture Imagery Program (NAIP) \cite{naip} collections used in this work) that is well-sampled across broad geographic domains via cloud-based platforms such as Google Earth Engine \cite{gorelick2017google}. 

Two broad ML techniques that have shown promising results using unlabeled image data are self-supervised learning and semi-supervised learning. Both these techniques involve training on unlabeled data, however, there are key differences in how unlabeled information is utilized.

\subsection{Self-supervised Learning}
Self-supervised learning techniques focus on learning image representations from unlabeled data by solving a specifically designed task that requires semantic understanding of the image but doesn't require manual labels. This \textit{pretext} task does not have to be similar to the eventual application, but solving it forces the model to learn representations that can encode prior knowledge of the input space (i.e. satellite imagery in our case). These representations can be used as feature extractors or can be fine-tuned for a specific downstream task with small amounts of labeled data. 
One category of self-supervised pretext tasks drops a part of the input signal and trains the network to predict it from the remaining information.
Such tasks include image in-painting (i.e. predicting an artificially cut out part of the image) \cite{pathakCVPR16context}, colorizing grayscale images \cite{zhang2016colorful}, predicting rotation angle \cite{gidaris2018unsupervised}, solving jigsaw puzzles (by predicting the correct order of shuffled patches) \cite{noroozi2016unsupervised} etc.
Models pre-trained on these tasks have shown impressive results on downstream tasks like natural image classification, object detection, and segmentation.
Self-supervision can also improve the robustness of the model against uncertainty and outliers \cite{hendrycks2019selfsupervised}.
Self-supervised methods have been adopted for various remote sensing tasks to learn better representations and improve transfer learning performance. 
For example, Zhao \textit{et al.} \cite{zhao2020self} train a model for scene-level (whole image) classification with an additional self-supervised task of rotation prediction.
Vincenzi \textit{et al.} \cite{vincenzi2020color} pre-train a model to predict natural-color (RGB) images from additional spectral bands (i.e. near- and short-wave infrared) and fine-tune it for land cover classification.
Lin \textit{et al.} \cite{lin2017marta} use GANs (Generative Adversarial Networks) with mid- and high- level feature fusion to learn image representations in an unsupervised manner for scene level land use classification and coffee plantation classification.
Lu \textit{et al.} \cite{lu2017remote} use unsupervised feature learning for scene classification from high spatial resolution imagery.
Jean \textit{et al.} \cite{jean2019tile2vec} pre-train their model with triplet loss to enforce similarity in the representations of neighboring image tiles.

More recently, contrastive self-supervised methods where the network is trained to embed the augmented versions of the same sample (positive pair) closer to each other while pushing away the embeddings of other samples (negative samples) have gained popularity.
Various contrastive self-supervised learning methods (e.g., SimCLR \cite{chen2020simple}, MoCo \cite{He_2020_CVPR}, BYOL \cite{grill2020bootstrap}, SwAV \cite{caron2020unsupervised}) have a similar setup and mainly differ in how they generate negative samples (with BYOL not using negative examples at all).
Representations learned by these contrastive models perform on par with the representations learned using purely supervised baselines on ImageNet classification \cite{ILSVRC15}, which involves classifying natural images into 1000 categories, and is a commonly used benchmark for ML methods. These representations also show state-of-the-art results for other vision tasks via transfer learning \cite{He_2020_CVPR, caron2020unsupervised}.
Such contrastive methods have also been explored and adopted for remote sensing tasks to improve performance on various benchmark datasets for image-level classification \cite{jung2021contrastive, stojnic2021self, ayush2020geography, kang2020deep}.
Leenstra \textit{et al.} \cite{leenstra2021self} train a self-supervised network to identify overlapping patches and make their representations similar to each other, which is then fine-tuned for a change detection segmentation task.
Li \textit{et al.} \cite{li2021semantic} train a multi-task self-supervised network using three pretext tasks (in-painting occluded patches, predicting relative transformation between two patches, and contrastive loss) and show better performance than ImageNet pre-trained baseline for fine-tuning on segmentation datasets.

\subsection{Semi-supervised Learning}
Unlike self-supervised learning, which uses only unlabeled data for learning and is defined independent of the labeled data and task, semi-supervised learning uses unlabeled data in conjunction with the labeled data and is closely tied to the actual supervised task. Semi-supervised learning methods work by encouraging the model to better generalize to unseen data, which is usually accomplished by adding an extra loss function. The loss function is often one of entropy minimization or consistency regularization. These loss functions improve model generalization by regularizing the model outputs on unlabeled data.
Entropy minimization encourages the model to output more confident class-predictions for the unlabeled data at the task of image classification, \cite{grandvalet2005semi} \cite{sajjadi2016mutual}.
Consistency regularization, on the other hand, encourages the network to output the same predictions for perturbations of the same input data point, and specific approaches include MixMatch \cite{berthelot2019mixmatch}, ReMixMatch \cite{berthelot2019remixmatch}, FixMatch \cite{sohn2020fixmatch}, and consistency-based object recognition \cite{jeong2019consistency}.
With these recent advancements from FixMatch \cite{sohn2020fixmatch} (and its successors, e.g. DivideMix \cite{li2020dividemix}),
semi-supervised learning has established itself as a promising technique for utilizing unlabeled data, especially in low label data regimes.

Semi-supervised techniques have historically been used with classical ML methods for a variety of remote sensing tasks, e.g. SVMs for image classification (\cite{camps2007semi}, \cite{tuia2009semisupervised}, \cite{bruzzone2006novel}) and segmentation (\cite{maulik2011self}, \cite{maulik2013learning}).
More recently, these tasks have been addressed using deep-learning methods like boundary-aware semantic segmentation of very-high ground-resolution aerial and satellite images (BASNet) \cite{sun2020bas}, individual tree canopy detection \cite{weinstein2019individual}, and scene classification from high-resolution optical satellite images \cite{han2018semi}.
Semi-supervised learning has also been used for retrieving images for a given semantic description of  land cover. For example, Chaudhuri \textit{et al.} \cite{chaudhuri2017multilabel} used a graph-theoretic method, while Tang \textit{et al.} \cite{tang2019large} utilized semi-supervised deep adversarial hashing (SDAH), and Hu \textit{et al.} \cite{hu2019mima} used semi-supervised manifold alignment (SSMA), a multi-modal data fusion algorithm for manifold alignment, while combining different data-sources like high-resolution optical, hyperspectral (simulated spaceborne EnMAP) and dual-polarimetric SAR (VV-VH polarized Sentinel-1).
Semi-supervised methods have also been used to address the problem of dimensionality reduction of hyperspectral data. For example, Hong \textit{et al.} \cite{hong2019learning} utilize iterative multitask regression framework and Wu \textit{et al.} \cite{wu2018semi} propose a semi-supervised method based on Local Fisher Discriminant Analysis.
More recent work by Hong \textit{et al.} \cite{hong2020x} explores semi-supervised cross-modal learning between high-resolution multi-spectral data (MSI), synthetic aperture radar (SAR) Sentinel-1 data, and low-resolution small-scale-hyperspectral data (HSI).
This cross-modal learning strategy enables their model to generalize well, and it achieves better performance than the state-of-the-art models at the task of classification.

\subsection{This study}
Given that self- and semi-supervised ML techniques are typically developed using standard three-channel (RGB) consumer camera imagery, there is a critical need to have a systematic study of these methods on Earth observation datasets, specifically with respect to label scarcity and geographic domain shifts. In this study, we test SimCLR\cite{chen2020simple}, a self-supervised method, and FixMatch\cite{sohn2020fixmatch}, a semi-supervised method, for segmentation tasks using the DeepLabv3+ neural network architecture\cite{Chen_2018_ECCV}. We consider three different tasks and types of remote sensing imagery in our evaluation. The first task involves segmenting riverbeds from high resolution (4m) RGB imagery, the second involves classifying land cover from high resolution (1m) RGB+NIR imagery from the United States Department of Agriculture National Agriculture Imagery Program (NAIP \cite{naip}) \cite{maxwell2017land}, and the third involves mapping flood extents from moderate resolution Synthetic Aperature Radar (SAR) imagery acquired by the European Space Agency's Sentinel-1 constellation \cite{torres2012gmes}.
We augment the labeled remote sensing datasets with unlabeled images sampled from the same imagery collections. With extensive experimentation, we analyze the improvements that SimCLR and FixMatch can provide by using this additional unlabeled imagery and study the impact of label scarcity and geographical domain-shifts.
We compare these methods against supervised fine-tuning from ImageNet pre-trained weights, a strong baseline not only for consumer-camera-based vision problems\cite{huh2016makes}\cite{kornblith2019better}, but also for remote sensing tasks like scene classification \cite{pires2020convolutional}.  We find that the remote sensing tasks considered here not only benefit from semi- and self-supervised training individually, but that combining these techniques allows us to reap even greater benefits when faced with limited labeled data and geographic domain shifts.

%% file: sections/methods.tex
We evaluated SimCLR and Fixmatch on the task of semantic image segmentation (i.e. providing a class label to each pixel in the image). We chose to work with neural network models as they are now the
state of the art on all computer vision benchmarks as well as many remote sensing tasks (especially those
that involve spatial context understanding and don't rely on just single pixel based features) \cite{ma2019deep, DLinRLComprehensive}.
The next section describes the supervised DeepLabv3+ \cite{Chen_2018_ECCV} segmentation model that is used across all our experiments, followed by an overview of the self-supervised representation learning model SimCLR\cite{chen2020simple} and semi-supervised learning model FixMatch \cite{sohn2020fixmatch}.

\subsection{Supervised DeepLabv3+}
\label{ssec:deeplab_supervised}
\begin{figure}
    \centering
    \includegraphics[width=0.9\linewidth]{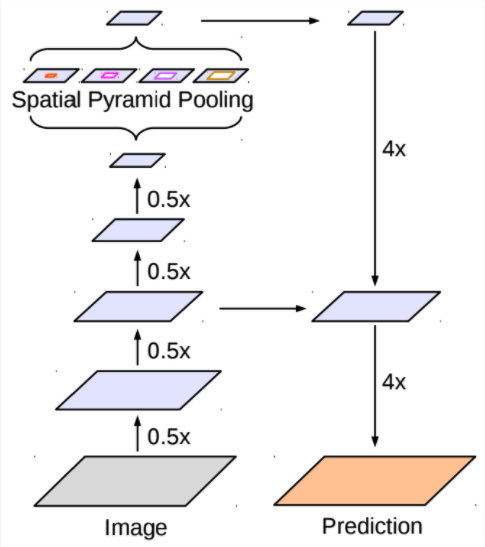}
    \caption{Encoder-decoder architecture of DeepLabv3+(Courtesy of \emph{Chen et al.} \cite{Chen_2018_ECCV}). A CNN feature encoder is followed by an atrous spatial pyramid pooling (ASPP) layer to learn features at different scales. A decoder with one skip connection upsamples the learned features before predicting the segmentation class probabilities of each pixel.}
    \label{fig:deeplab}
\end{figure}

DeepLabv3+\cite{chen2020simple} is a widely used convolutional neural network (CNN) architecture for semantic image segmentation which has shown state-of-the-art results on several benchmark datasets for object detection and segmentation (e.g., MSCOCO \cite{Lin2014MicrosoftCC}, PASCAL VOC \cite{Everingham10}).
In traditional encoder-decoder based segmentation models like SegNet \cite{Badrinarayanan17}, max-pooling and/or striding in the encoder CNN reduces the spatial resolution of feature maps, which are then upsampled by deconvolution layers in the decoder.
DeepLabv3+ uses a similar architecture but utilizes atrous convolutions \cite{Chen_2018_ECCV} in its backbone encoder to learn deep features without degrading spatial resolution.
These feature maps are further processed by ASPP (atrous spatial pyramid pooling) to learn features at different scales. The decoder consists of an upsampling plus a convolutional layer with a skip connection to the corresponding encoder layer.
Figure \ref{fig:deeplab} illustrates the high level model architecture for Deeplab.
Final feature maps are upsampled to the original input size and projected to output logits, and cross entropy loss is computed for each pixel against the ground truth pixel label.
For more technical details, please refer to \cite{Chen_2018_ECCV}.

\subsubsection*{Augmentations}
\label{sssec:deeplab_aug}
Input augmentation is a standard practice to prevent overfitting, regularize models, and improve generalization \cite{perez2017effectiveness} \cite{shorten2019survey}, and many semi- and self- supervised methods rely on augmentations of input images as part of their learning process.
Hence, it was important to ensure that the augmentations used in our baseline model were thoroughly optimized to avoid conflating the benefits of self- and semi-supervised techniques with the augmentation improvements associated with them.
The final set of optimized augmentations used for supervised DeepLabv3+ learning (and fine-tuning experiments) is described below.
See the \ref{sec:appendix}{Appendix} for more details and ablation on these augmentations.
\begin{itemize}
    \item \emph{Random crop with distortion}: We take a random crop of predefined input size from the dataset image. Crop distortion is set to $s=0.5$ which means that each side of the image is randomly stretched between $(1+s)$ and $(1-s)$ times the original size. This augmentation makes the model robust to minor perturbations in resolution.
    \item \emph{Rotation/flips}: Each cropped image undergoes random horizontal flip, random vertical flip and random 90-degree rotations, making the model robust to these simple geometric transformations.
    \item \emph{Appearance augmentation}: Color jitter is applied with 0.5 probability. Color jitter randomly changes brightness, contrast, saturation, and hue of an RGB image. However, saturation and hue augmentations are not meaningful for non-RGB images (such as RGB+NIR and SAR images). For such images, we replace saturation and hue with per channel brightness and contrast augmentations. As opposed to standard contrast and brightness which apply augmentation on each channel by the same strength factor, per-channel brightness and contrast are applied on each channel with a separate randomly chosen factor. This generalized color jitter, which includes brightness, contrast, per-channel brightness and per-channel contrast, makes our model robust to variations coming from the imaging sensor, post-processing, and the domain of training data, and forces it to consider spatial structures over color when making predictions.
\end{itemize}

\subsection{SimCLR Self-supervised model}
\begin{figure}
    \centering
    \includegraphics[width=0.9\linewidth]{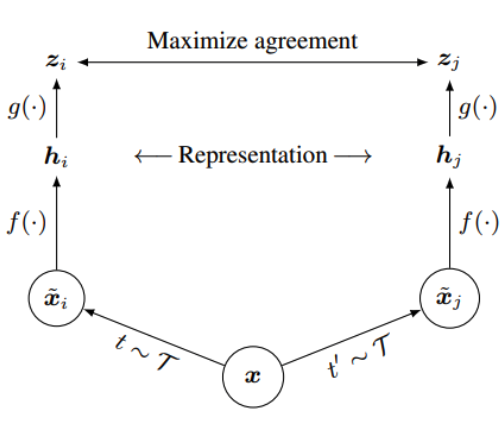}
    \caption{SimCLR framework (Courtesy of \cite{chen2020simple}).}
    \label{fig:simclr}
\end{figure}
\begin{figure*}[htb]
 \center
  \includegraphics[width=0.9\textwidth]{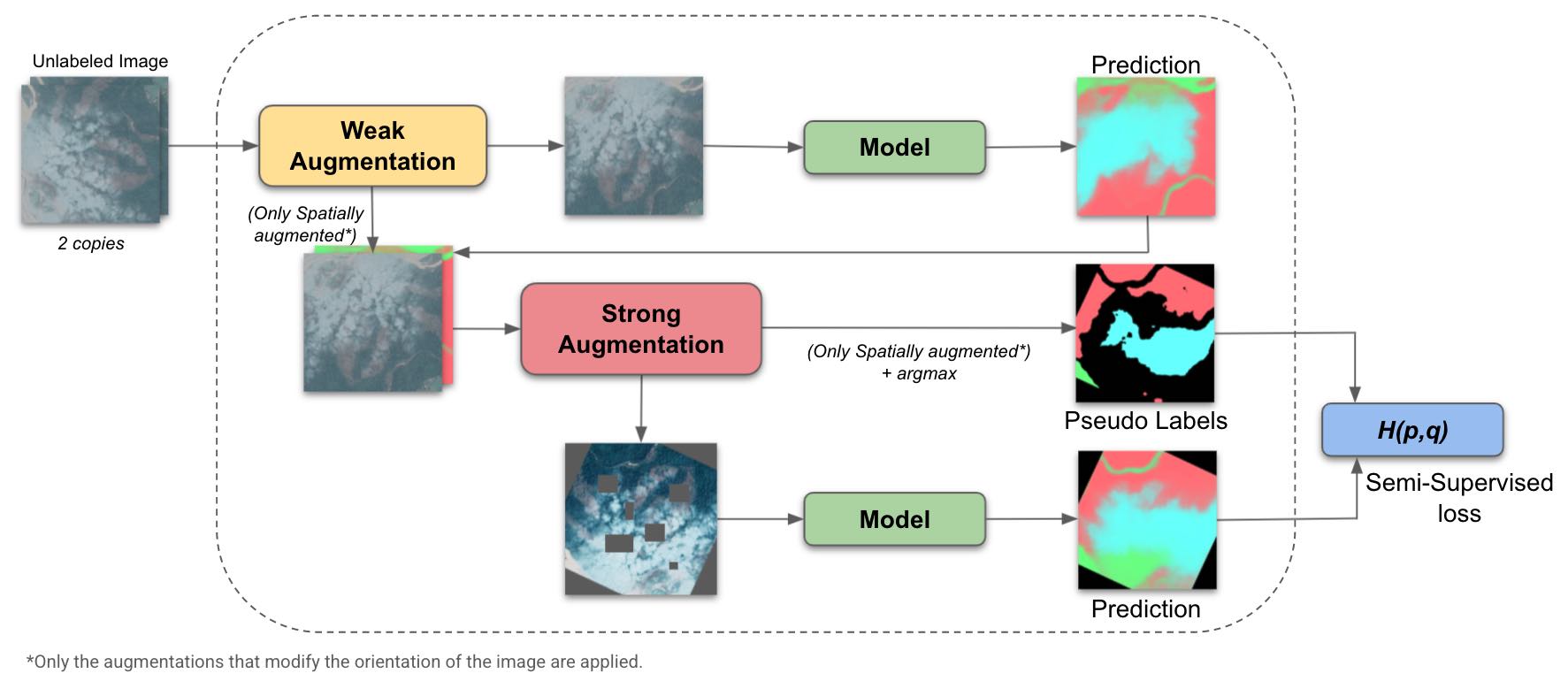}
  \caption{FixMatch architecture with the semi-supervised loss adopted for semantic segmentation. The model generates predictions for weakly augmented unlabeled images. Weakly augmented images and their predictions undergo strong augmentations, with only the spatial augmentations being applied to predictions to generate pseudo labels. Semi-supervised cross-entropy loss is computed on this strongly augmented image and pseudo labels.}
  \label{fig:fixmatch}
\end{figure*}

SimCLR \cite{chen2020simple} is a self-supervised learning method that maximizes the similarity between representations of differently augmented views of the same image by applying a contrastive loss \cite{hadsell2006dimensionality}. Models pre-trained using SimCLR provide a much better network initialization than random weights for downstream tasks, especially when labeled data is scarce.
A ResNet-50 \cite{he2016deep} based SimCLR model \cite{chen2020simple} trained only on ImageNet images (without labels) achieves 48.3\% top-1 accuracy when fine-tuned over 1\% of labeled ImageNet data, surpassing the supervised baseline (with random initialization), which achieves only 25.4\% top-1 accuracy.

The contrastive loss is defined per mini-batch of images. Specifically, given a randomly sampled mini-batch containing N examples, each image is augmented twice, resulting in $2N$ augmented images $\{x_{\emph{k}}, k \in [1, ..., 2N]\}$.
Each augmented image $\{x_{\emph{i}}\}$ has a corresponding positive example $\{x_{\emph{j}}\}$ and the remaining $2(N-1)$ images are considered negative examples.
These images are encoded with the encoder network $f(.)$ to generate feature representations $\{h_{\emph{k}}\}$ (see Figure \ref{fig:simclr}), and we use the same encoder used in DeepLabv3+ in order to allow transfer learning.
The resulting feature representations are then fed to a non-linear projection network $g(.)$ to generate $\{z_{\emph{k}}\}$.
For each positive pair $(i, j)$ of the mini-batch, SimCLR computes the contrastive loss as follows:

\begin{equation*}
    l_{i,j} = -\log\frac{
    exp(sim(z_i,z_j)/\tau)
    }{
    \sum_{k=1}^{2N} \mathds{1}_{[k\neq i]} exp(sim(z_i,z_k)/\tau)
    }
\end{equation*}

\noindent where $sim(z_i,z_j)$ denotes the cosine similarity between vectors $z_i$ and $z_j$. 
$1_{[k \neq i]} \in \{0, 1\}$ is an indicator function evaluating to $1$ \emph{iff} $k \neq i$  and $\tau$ denotes a temperature parameter. 
Intuitively, this loss maximizes the similarity between $z_{\emph{i}}$ and $z_{\emph{j}}$ (coming from the augmented views of the same image) and minimizes their similarities to other $\{z_{\emph{k}}, k \notin \{i, j\} \}$ coming from different images from the same minibatch.

\subsubsection*{Augmentations}
To generate augmented views of the input image, the original SimCLR implementation \cite{chen2020simple} uses random distorted crop, random horizontal flip, random color distortion (color jitter and color dropping), and random Gaussian blur.
We use the same augmentation policy for our work.
However for non-RGB inputs (such as RGB+NIR and SAR images), we use general color jitter as explained in Section \ref{sssec:deeplab_aug}. Further details on the specific augmentation parameters are provided in the Appendix.

To leverage SimCLR self-supervised training on downstream segmentation tasks, we initialize the encoder of the DeepLabv3+ segmentation network with a trained SimCLR encoder $f(.)$ and fine-tune on the labeled dataset.
The remaining parts of DeepLabv3+ architecture like ASPP layers and decoder layers, which constitute a small fraction of total parameters, are trained from scratch with random initialization.
For example in the Resnet-50 based DeepLabv3+ model, 24M encoder parameters are initialized with SimCLR pre-trained weights and remaining 3M parameters are trained from scratch.

\subsection{FixMatch Semi-supervised model}

FixMatch \cite{sohn2020fixmatch} is a label-propagation based semi-supervised learning algorithm for image classification.
It has shown state-of-the-art performance in extremely low-data regimes across a variety of standard image classification benchmarks, including 94.93\% accuracy on CIFAR-10 \cite{krizhevsky2009learning} (a 10-way classification task) with 250 labeled images and 88.61\% accuracy with 40 labeled images (i.e four labels per class).
The key idea behind FixMatch is to use unlabeled data for consistency regularization (\cite{jeong2019consistency,sajjadi2016regularization}).
Each unlabeled image undergoes two types of augmentations: weak and strong.
Weak augmentations refer to less deforming augmentations like horizontal/vertical flipping, color jitter etc., and strong augmentations refer to more deforming augmentations like shear (a stronger spatial transformation), rotation, solarization (inverting colors above a threshold), and posterization (reducing pixel bits).
Pseudo-labels are generated using the model predictions with current weights on weak augmentations of unlabeled images.
Only predictions that pass a threshold for the predicted probability (implying a high model confidence) are used as pseudo-labels. These pseudo-labels are then used as training labels for stronger augmentations of the same input image via addition of an additional cross-entropy loss, i.e., the semi-supervised loss.
The total loss for the model is a weighted average of supervised loss (using labeled samples), and semi-supervised loss (using unlabeled samples).
By considering only pseudo-labels that are above a certain confidence threshold, FixMatch avoids the careful balancing of these losses needed by MixMatch \cite{berthelot2019mixmatch} and ReMixMatch \cite{berthelot2019remixmatch}, \cite{tarvainen2018mean}, and \cite{oliver2019realistic}.
The maximum probability across all classes is usually low during the beginning of the training and increases as the training progresses, producing a curriculum effect in training. This enables training of Fixmatch on datasets with as few as a single labeled sample per class.

We extend the FixMatch algorithm from scene-level classification to pixel-wise segmentation tasks. Unlabeled images are passed through the DeepLabv3+ model to generate per pixel pseudo-labels and valid masks are computed by thresholding the resulting probabilities. The prediction (and the validity mask) on the weakly augmented image are then spatially transformed to generate pseudo-labels that are correctly aligned with the strongly augmented image (see Figure \ref{fig:fixmatch}).

\subsubsection*{Augmentations}
The same augmentations used for supervised DeepLabv3+ learning (rotation/flipping and color jitter, see Section \ref{ssec:deeplab_supervised}) are used as weak augmentations for FixMatch.
For strong augmentations, we use the same configuration for the strong augmentations proposed by Cubuk \textit{et al.} in Autoaugment \cite{cubuk2019autoaugment}: out of a list of many augmentation functions (like equalize, solarize, posterize, shear, and rotation), two augmentations are chosen and applied in succession, followed by a cutout, which removes a part of the image. However, unlike FixMatch, which removes a single large rectangle for cutout, we remove multiple smaller rectangles.
The complete details of the augmentations can be found in the \ref{sec:appendix}{Appendix}.

%% file: sections/datasets.tex
We selected tasks and datasets for model evaluation based on the following criteria:
(i) The task involves pixel-wise segmentation, where each pixel in the image is assigned a label,
(ii) We have the ability to sample unlabeled imagery outside of the original dataset from the same imaging platform/source, and
(iii) Imagery is georeferenced and the dataset can be split into geographically distinct, non-overlapping training, validation, and test regions.

For each dataset, we create two types of train/validation/test splits in order to test for domain generalization:
(i) an IID (Independent and Identically Distributed) partition, where the train/validation/test splits are created by IID sampling the entire dataset, and
(ii) a domain-shifted partition, where the train/validation/test splits are sampled from different geographical areas.

The next three subsections give specific details on the segmentation tasks and datasets evaluated in this work, and Table \ref{tab:dataset_details} provides a general overview of key attributes.

\subsection{Riverbed Segmentation Dataset}

\begin{figure}[htb]
    \centering
    \includegraphics[width=\linewidth]{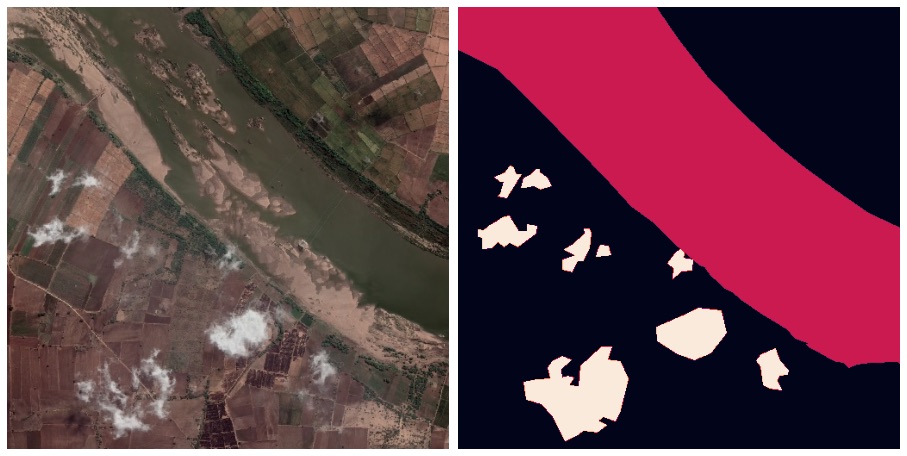}
    \caption{Riverbed segmentation task example with the RGB image on the left and the segmentation ground-truth labels on the right, with the riverbed shown in red, clouds in white and background in black. Note that riverbeds are defined as including both the water and the sandy portion of the river.}
    \label{fig:riverbed_example}
\end{figure}
\begin{figure}[htb]
    \centering
    \includegraphics[width=0.47\linewidth]{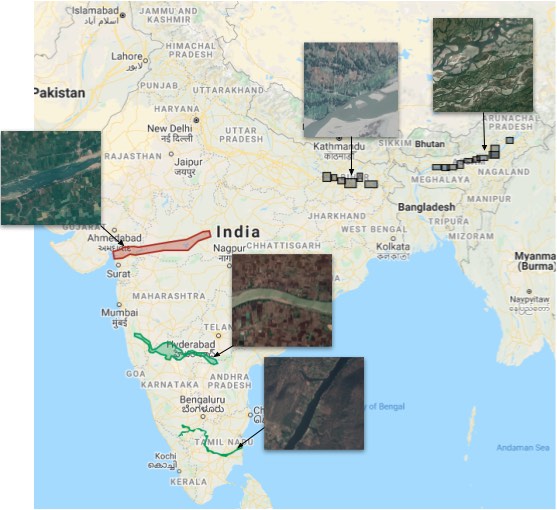}
    \includegraphics[width=0.45\linewidth]{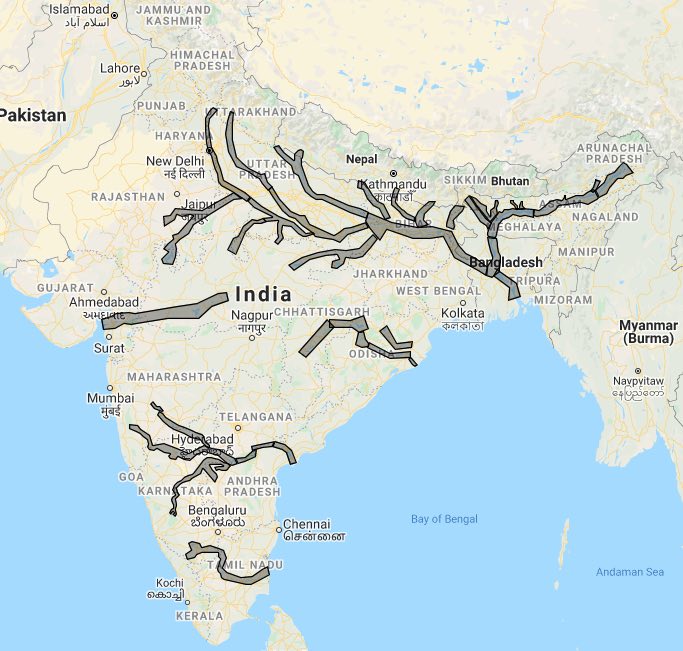}
    \caption{(Left) The riverbed regions where satellite images were sampled for the labeled dataset. As shown in the samples, each riverbed has distinct characteristics. (Right) The regions where the unlabeled dataset was sampled.}
    \label{fig:riverbed_sampling}
\end{figure}

The riverbed segmentation dataset was developed to demarcate riverbeds (i.e. both the water and sandy portion of a river channel) from high-resolution RGB images sampled from Google Maps basemap imagery (see Figure \ref{fig:riverbed_example}).
This custom dataset consists of 26,112 RGB images of size 513x513 at 4m resolution sampled from riverbed regions of five Indian rivers: Ganga, Brahmaputra, Narmada, Krishna, and Kaveri (see Figure \ref{fig:riverbed_sampling}).
Each image was labeled by a single human interpreter who classified each pixel into one of the three classes -- riverbed, clouds, and background -- using a polygon drawing tool.

\begin{figure*}
    \centering
    \includegraphics[width=\textwidth]{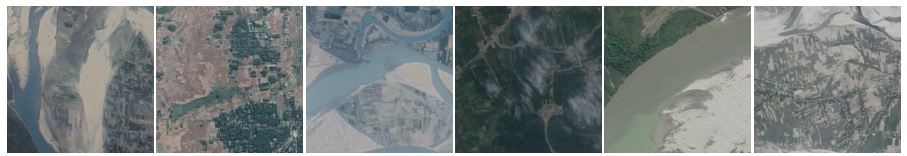}
    \includegraphics[width=\textwidth]{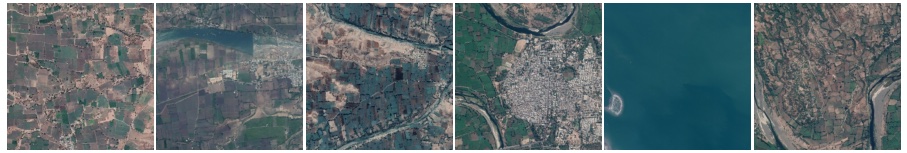}
    \includegraphics[width=\textwidth]{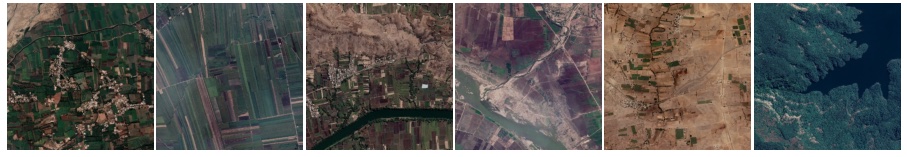}
    \caption{Randomly sampled images from train (top), validation (middle) and test (bottom) split of domain-shifted partition of Riverbed segmentation dataset. It can be noticed that the riverbed differs in structure and color among the splits.}
    \label{fig:riverbed_domain_shift}
\end{figure*}

The IID partition of this dataset was done using a 60:20:20 ratio, resulting in 15,708 training, 5,243 validation, and 5,161 test images.
The domain-shifted partition sampled 15,504 training images from the Ganga and Brahmaputra riverbed regions (black polygons in Figure \ref{fig:riverbed_sampling}(Left)), 7,568 validation images from the Narmada riverbed (red polygon in Figure \ref{fig:riverbed_sampling}(Left)), and 3,040 test images from the Krishna and Kaveri riverbeds (green polygon in Figure \ref{fig:riverbed_sampling}(Left)).
Each riverbed has distinct visual characteristics, thus making it harder for models trained on one region to generalize on the other. See Figure \ref{fig:riverbed_domain_shift} for example imagery.

The unlabeled data around riverbed regions of several rivers of India (polygons in Figure \ref{fig:riverbed_sampling}(Right)) was sampled from the same image collection as the labeled data.
In total, an additional 183,668 RGB images of size 1024x1024 were sampled at the same 4m resolution. We exhaustively sampled all available imagery from the polygons in Figure \ref{fig:riverbed_sampling}(Right) -- as the polygons used for unlabeled imagery were a superset of the polygons used for labeled imagery, the unlabeled images included all the images from the labeled set. The same unlabeled data was used for both semi-supervised and self-supervised experiments.

\subsection{Chesapeake Land Cover}

\begin{figure}[htb]
    \centering
    \includegraphics[width=\linewidth]{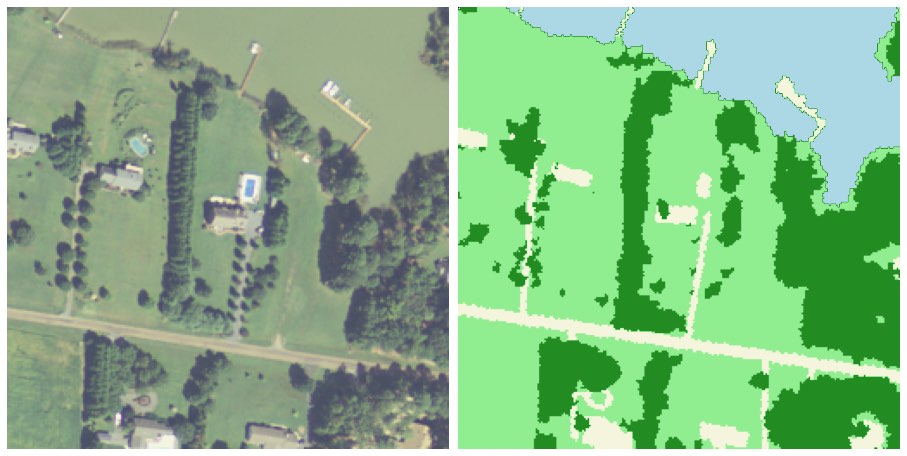}
    \caption{Chesapeake Land Cover task example with the RGB image on the left and the segmentation groundtruth label on the right. [Blue: Water, Light green: Low vegetation, Dark green: Forest, Light yellow: Impervious land]}
    \label{fig:chesapeake_task}
\end{figure}

\begin{figure}[htb]
    \centering
    \includegraphics[width=\linewidth]{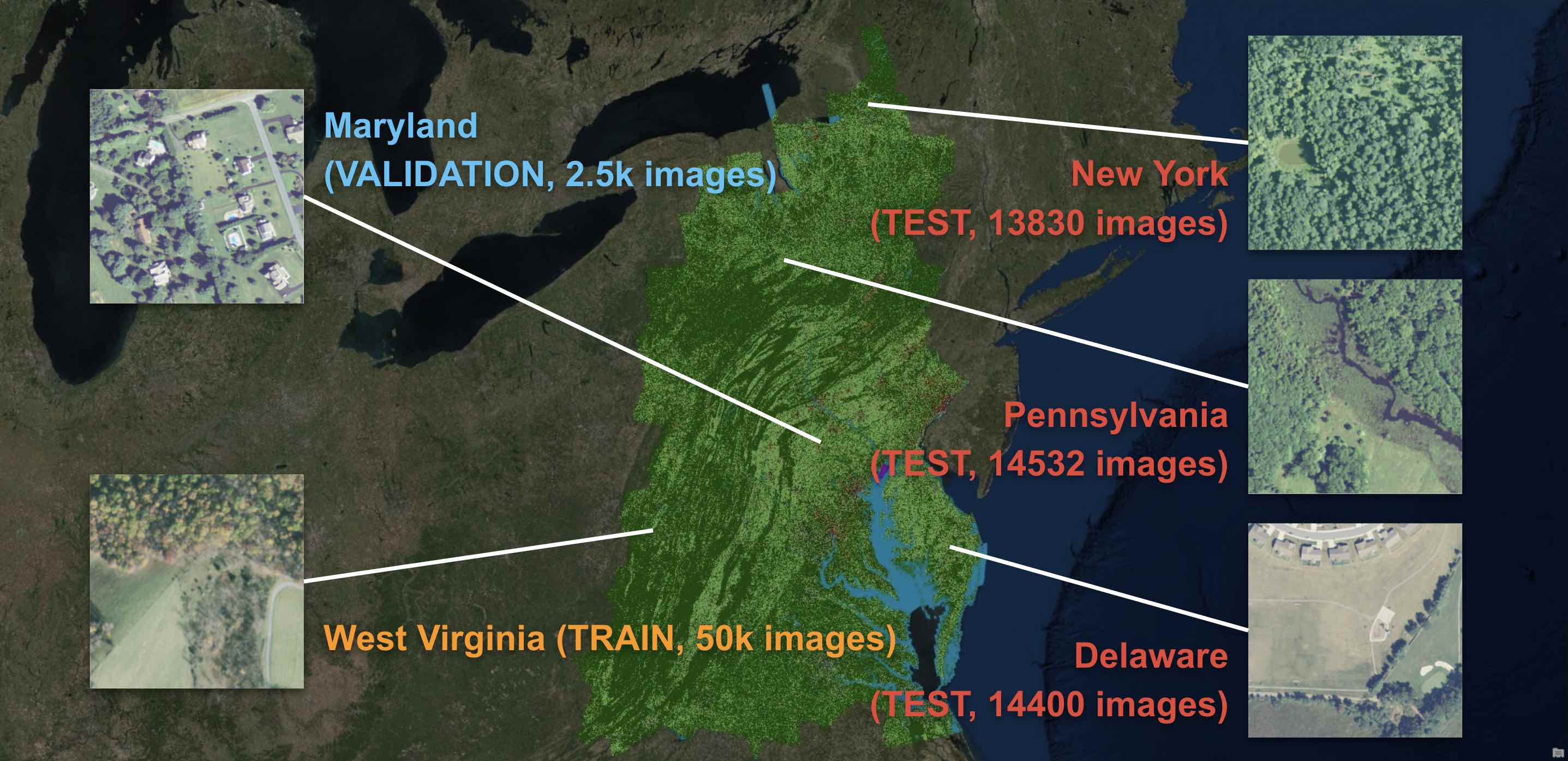}
    \caption{The region where Chesapeake Land Cover data was sampled, along with the domain-shifted partition and sample images from each state.}
    \label{fig:chesapeake_sampling_labeled}
\end{figure}

\begin{figure}[htb]
    \centering
    \includegraphics[width=0.45\linewidth]{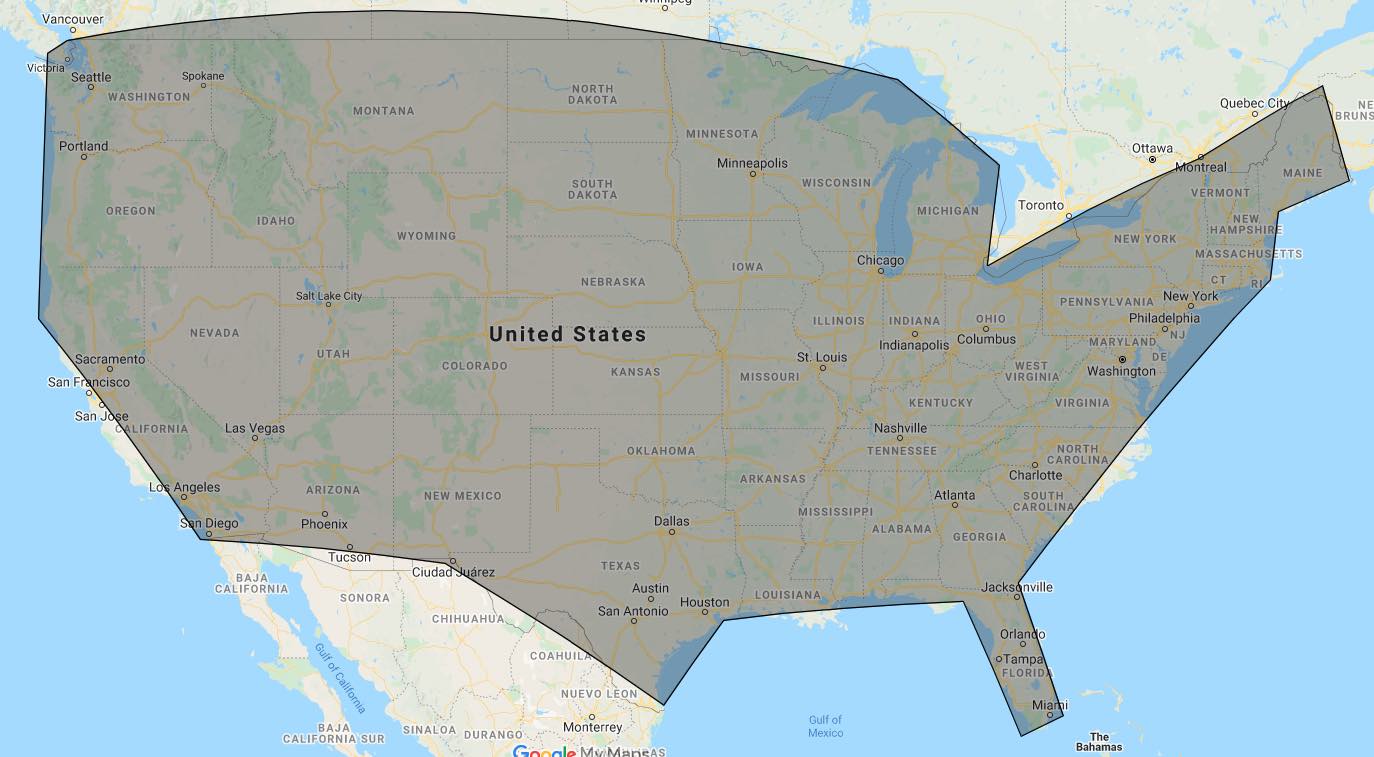}
    \includegraphics[width=0.45\linewidth]{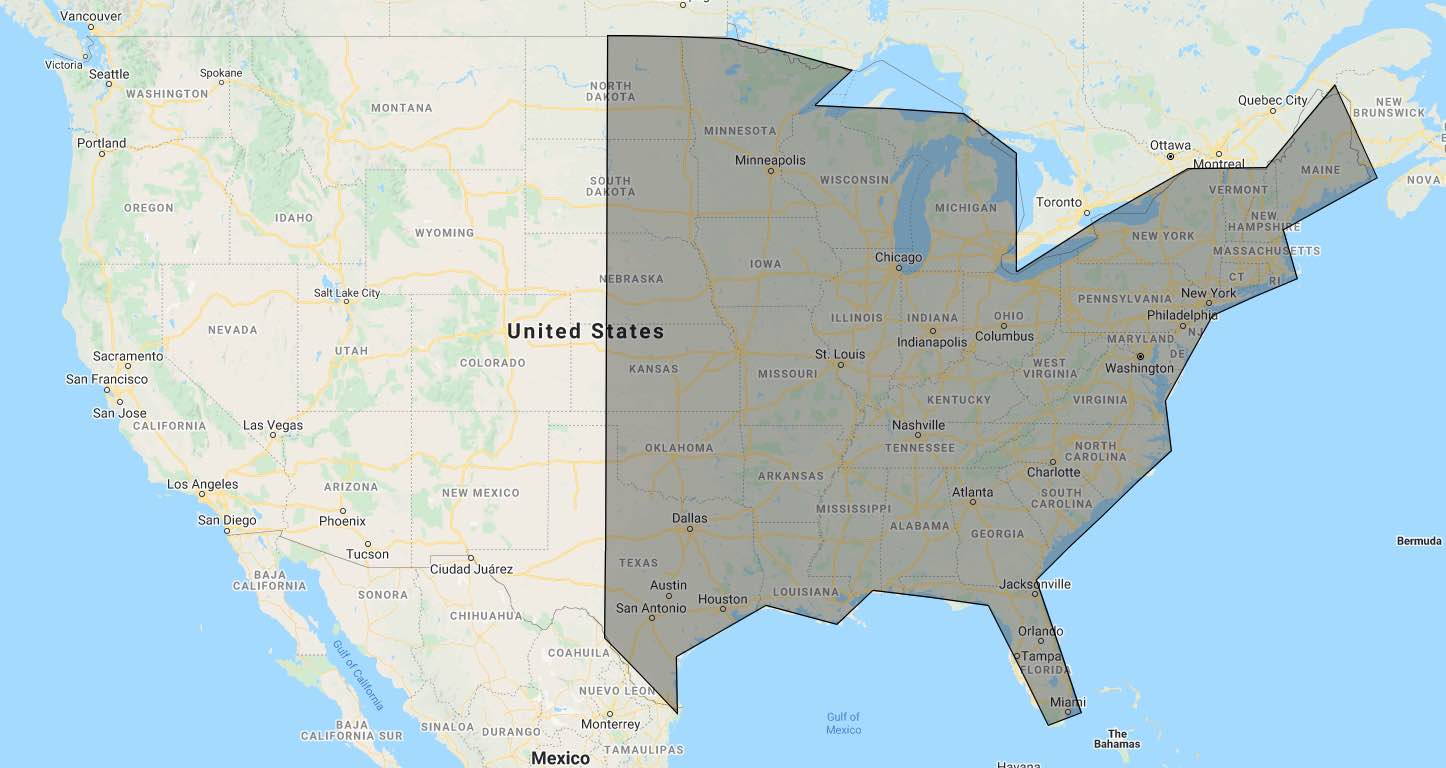}
    \caption{(Left) Unlabeled data from the mainland USA was used to train the SimCLR model for Chesapeake dataset. (Right) Unlabeled data from the east half of the mainland USA was used to train FixMatch semi-supervised models.}
    \label{fig:chesapeake_unlabeled}
\end{figure}

The publicly available Chesapeake Land Cover dataset\cite{robinson2019large} is sourced from the Chesapeake conservancy region in the eastern United States.
It covers an area of about 100,000 sq miles that spans over six states: New York, Pennsylvania, Maryland, Delaware, Virginia, and West Virginia (Figure \ref{fig:chesapeake_sampling_labeled}).
The dataset consists of multi-spectral imagery from the NAIP program\cite{naip} at 1m resolution.
Labels were created using semi-automated feature extraction and rule-based clustering, followed by corrections from experts. The complete details can be found on their website \cite{chesa_land_cover_data_project_2020}.
Pixels are classified into four categories: (i) water, (ii) forest, (iii) field, and (iv) impervious surfaces (Figure \ref{fig:chesapeake_task}).
For our experiments, we use the visible (RGB) and the near-infrared bands (NIR) for images of size 256x256 pre-sampled by the dataset creators.
Each state has a pre-populated IID split of 50,000 training samples and 2,500 validation samples per state.
The test split is provided as large TIFF tiles ($\sim$ 6000x7500), which we slice into images of size 256x256.

\begin{figure*}
    \centering
    \includegraphics[width=\textwidth]{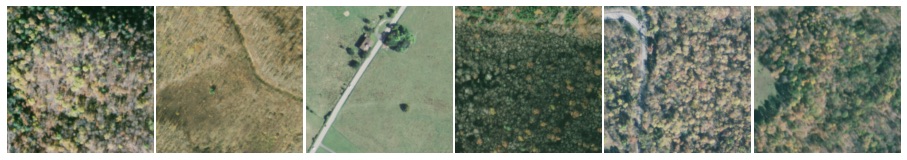}
    \includegraphics[width=\textwidth]{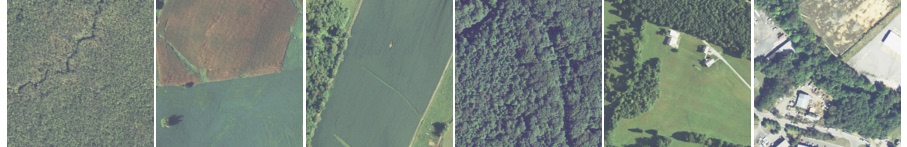}
    \includegraphics[width=\textwidth]{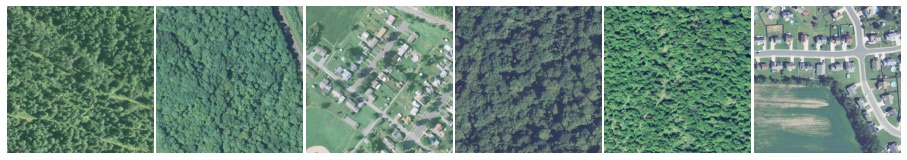}
    \caption{Randomly selected images from train (top), validation (middle) and test (bottom) split of the domain-shifted partition of the Chesapeake Land Cover dataset.}
    \label{fig:chesapeake_domain_partition}
\end{figure*}

For the IID partition, we use the data from the state of Maryland, consisting of 50K training examples, 2,500 validation examples and 14,750 testing examples selected from 20 test tiles from the same state. For the domain-shifted partition, the training split consists of the training data from West Virginia (50k examples), the validation split uses the validation data from Maryland (2.5k examples), and the test split uses the test data from New York, Pennsylvania and Delaware (13,830+14,532+14,400=42,762 examples from a total of 60 tiles). See Figure \ref{fig:chesapeake_domain_partition} for example imagery.

Unlabeled data for this task was sampled from publicly available NAIP imagery (RGB+NIR) using Google Earth Engine \cite{naipee}. For SimCLR training, approximately 2 million 256x256 sized images were uniformly drawn from the entire region of the mainland United States from 2011 to 2014 (see Figure \ref{fig:chesapeake_unlabeled}(left)). This dataset contains a rich variety of geological features that is useful in representation learning for the SimCLR model. A second sample of unlabeled data containing approximately 2 million images was drawn from the eastern half of the United States (see Figure \ref{fig:chesapeake_unlabeled}(right)) for the FixMatch model.

\subsection{Sen1Floods11}
\begin{figure}[htb]
    \centering
    \includegraphics[width=\linewidth]{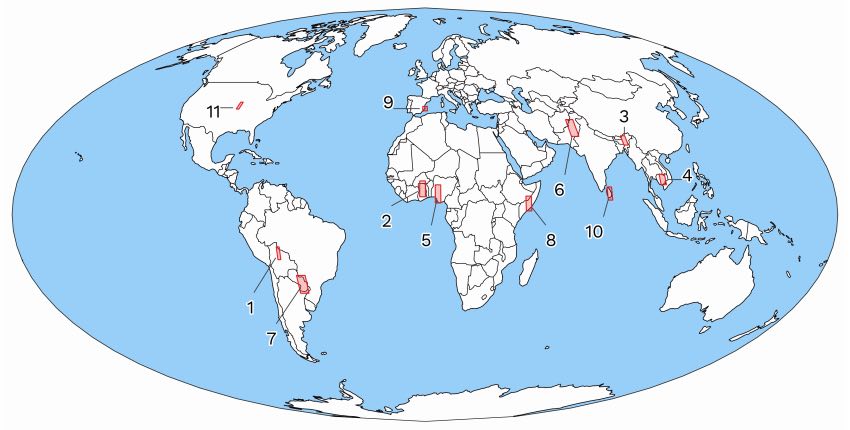}
    \caption{The locations of 11 flood events where the flood data of Sen1Floods11 was sampled (Courtesy of \cite{Bonafilia_2020_CVPR_Workshops}).}
    \label{fig:sen1floods11_events}
\end{figure}
\begin{figure}[htb]
    \centering
    \includegraphics[width=0.32\linewidth]{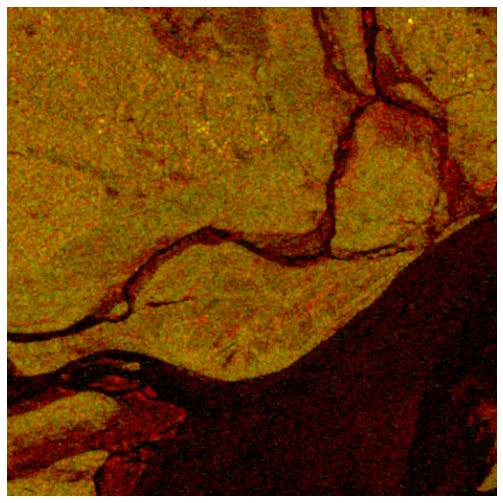}
    \includegraphics[width=0.32\linewidth]{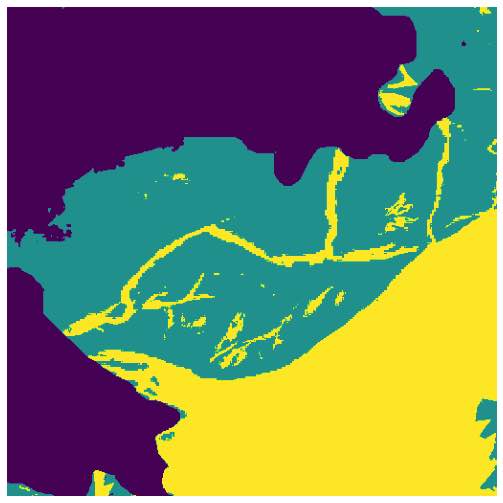}
    \includegraphics[width=0.32\linewidth]{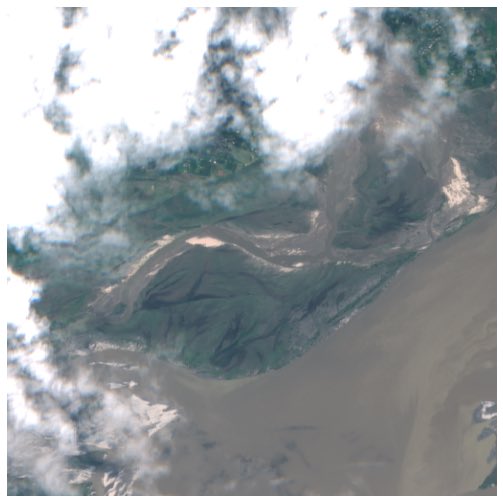}
    \caption{Sentinel-1 image (left), groundtruth floodmap (middle) and Sentinel-2 image (right) from Sen1Floods11 dataset. Sentinel-2 images are only for visualization and not used during training. Labels include water (yellow), no-water (cyan), and masked out region (purple).}
    \label{fig:sen1floods11_example}
\end{figure}

\begin{table*}[hb]
    \centering
    \ra{1.2}
    \begin{tabular}{@{} l l l l @{}}
    \toprule
     & Riverbed Segmentation & Chesapeake Land Cover & Sen1Floods11 \\
    \midrule
    Source & Google Earth & NAIP & Sentinel-1 \\
    Bands & RGB & RGB+NIR & VV and VH \\
    Resolution & 4m & 1m & 10m \\
    Image size & 513x513 & 256x256 & 512x512 \\
    Label classes & 3 (Riverbed, Clouds, Background) & 4 (Water, Forest, Field, Impervious) & 2 (Water, No-water) \\
    \addlinespace[2ex]
    Domain-shifted Partition \\
    \midrule
    No. Train images & 15,504 & 50,000 & 275 \\
    No. Validation images & 7568 & 2500 & 83 \\
    No. Test images & 3040 & 42,762 & 88 \\
    Train regions & Ganga, Brahmaputra & West Virginia & Ghana, India, Nigeria, Paraguay, USA \\
    Validation regions & Narmada & Maryland & Somalia, Sri-Lanka, Bolivia \\
    Test regions & Krishna, Kaveri & New York, Pennsylvania, Delaware & Mekong, Pakistan, Spain \\
    \addlinespace[2ex]
    IID Partition \\
    \midrule
    No. Train images & 15,708 & 50,000 & 252 \\
    No. Validation images & 5243 & 2500 & 89 \\
    No. Test images & 5161 & 14,750 & 90 \\
    \addlinespace[3ex]
    Unlabeled Dataset for FixMatch \\
    \midrule
    Source & Google Earth & NAIP & Sentinel-1 \\
    Image size & 1024x1024 & 256x256 & 512x512 \\
    No. Images & 183,668 & 1,999,840 & 4384 \\
    Sampling regions & All major rivers of India & East half of mainland USA & 11 flood events from Sen1Floods11 \\
    \addlinespace[2ex]
    Unlabeled Dataset for SimCLR \\
    \midrule
    Source & \multirow{4}{*}{Same as Above} & NAIP & Sentinel-1 \\
    Image size &  & 256x256 & 512x512 \\
    No. Images &  & 1,999,675 & 4384(above) + 63k \\
    Sampling regions &  & Mainland USA & Global \\
    \bottomrule
    \end{tabular}
    \caption{Summary of key attributes for the three datasets used for evaluation. The first section outlines information about the labeled dataset and image sources, followed by details about the domain-shifted partition, IID partition and unlabeled dataset sampling for FixMatch and SimCLR.}
    \label{tab:dataset_details}
\end{table*}

\begin{figure*}
    \centering
    \includegraphics[width=\textwidth]{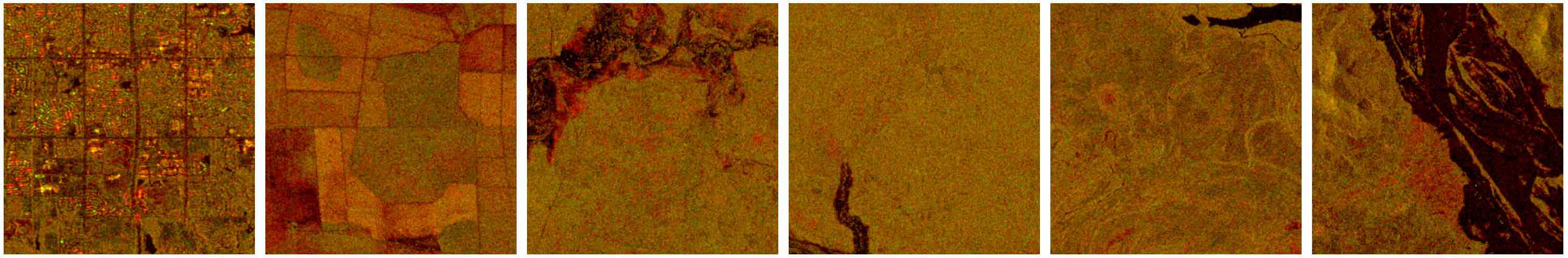}
    \includegraphics[width=\textwidth]{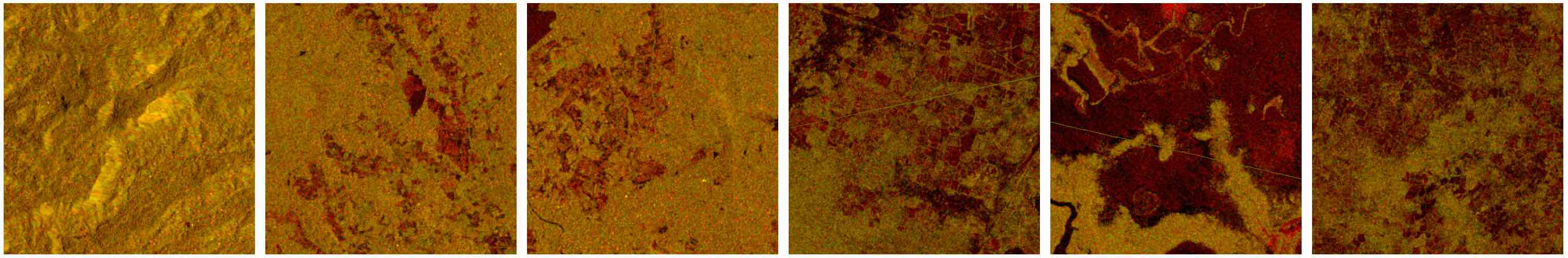}
    \includegraphics[width=\textwidth]{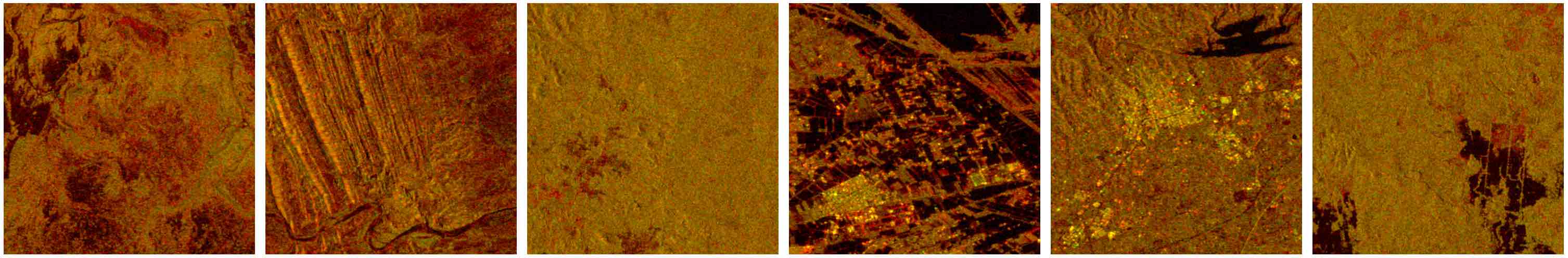}
    \caption{Randomly selected Sentinel-1 images from train (top), validation (middle) and test (bottom) split of domain-shifted partition of Sen1Floods11 dataset.}
    \label{fig:sen1floods11_domain_shift}
\end{figure*}

Sen1Floods11 \cite{Bonafilia_2020_CVPR_Workshops} is a surface water dataset that pairs raw satellite imagery with classified permanent and flood water.
The dataset includes 4,831 Sentinel-1 SAR images (Level-1 Ground Range Detected products, VV and VH polarizations) and aligned Sentinel-2 multi-spectral images (including all 13 spectral bands) sampled during 11 flood events from different regions of the world (see Figure \ref{fig:sen1floods11_events}).
The dataset authors created flood maps using Sentinel-2 images by thresholding NDVI (Normalized Difference Vegetation Index) and MNDWI (Modified Normalized Difference Water Index) values.
Out of the 4,831 images in the full dataset, 446 were then selected by the authors to be hand corrected by experts yielding high quality water labels. The final task is to perform flood segmentation using Sentinel-1 SAR images only (see Figure \ref{fig:sen1floods11_example}).

We carry out all of our experiments on the 446 expert-labeled images because they have high quality labels (curated by experts) and because this allows us to test self- and semi-supervised methods under the real-world constraint of an extremely small dataset.
The authors provide an IID partition of the Sen1Floods11 dataset containing 252 training samples, 89 validation samples, and 90 test samples (excluding the Bolivia flood event).
The same partition was used as the IID partition for our experiments. Our domain-shifted partition includes 275 training images from five locations (Ghana, India, Nigeria, Paraguay, and USA), 83 validation images from three locations (Somalia, Sri Lanka, Bolivia) and 88 test images from the remaining three locations (Mekong, Pakistan, Spain). See Figure \ref{fig:sen1floods11_domain_shift} for example imagery.

The 4,385 images that were not labeled by experts are used as unlabeled data for FixMatch semi-supervised learning (weak labels are not used).
However, this unlabeled dataset is quite small for SimCLR to learn rich features.
Therefore, we sample additional ~63K Sentinel-1 images randomly distributed across the globe from the year 2019. This large dataset, combined with 4,385 unlabeled images from Sen1Floods11, is used to train the SimCLR model.

%% file: sections/experiments_results.tex
The next section describes the overall experiment design, followed by a separate section each for dataset-specific setup and its results.

\subsection{Experiment Design}

For each dataset, we compare the performance of baseline (supervised) models that only use labeled data and the models that use unlabeled data. Refer to Table \ref{tab:model_comparison} for the list of models that are compared.

\begin{table*}[htb]
    \centering
    \ra{1.2}
    \begin{tabular}{@{} l l l l @{}}
    \toprule
    Model & Learning Method & Encoder initialization & Uses unlabelled data \\
    \midrule
    \suprandom & Supervised cross-entropy minimization & Random & No \\
    \supimgnet & Supervised cross-entropy minimization & ImageNet classification pre-training & No \\
    \midrule[\lightrulewidth]
    \supsimclr & Supervised cross-entropy minimization & SimCLR pre-training & Yes \\
    \fmrandom  & Semi-Supervised FixMatch & Random & Yes \\
    \fmimgnet  & Semi-Supervised FixMatch & ImageNet classification pre-training & Yes \\
    \fmsimclr  & Semi-Supervised FixMatch & SimCLR pre-training & Yes \\

    \bottomrule
    \end{tabular}
    \caption{List of models used in our study. First two models are baselines that only use labeled data. Next three models use unlabeled data for either pre-training or training. The last model (i.e. \fmsimclr) combines the benefit of both self and semi-supervised learning. All models use DeepLabv3+ architecture. Decoder is always initialized randomly.}
    \label{tab:model_comparison}
\end{table*}

To understand how performance varies with label scarcity, we sub-sample the labeled part of each dataset at 1\%, 10\% and 100\%.
We sample five draws of 1\% of the dataset and three draws for the 10\% dataset. These samples are chosen only once, do not overlap with each other and are fixed for all experiments.
However, for Sen1Floods11, 1\% of the labeled data amounts to only three labeled images which is too small a sample for supervised learning/fine-tuning.
Hence, we conduct our experiments only on 100\% and five samples of 10\% for this dataset.
The results on these partial datasets are reported by mean and standard deviation of the chosen metric.
We run the 100\% experiment three times as well to measure the variance in model training and report aggregated metrics similarly.

All models are trained and evaluated on both the IID and domain-shifted partition of each dataset.
The domain-shifted partition often reflects real-world deployment scenarios and provides a test bed for quantifying improvements that result from using unlabeled data, as this data provides an opportunity for the network to learn to generalize better.
For each dataset partition, the neural network weights are optimized on the training split, while all hyperparameters selection, checkpoint selection and experimental analysis is done on the validation split.
Once we have frozen our model and all hyperparameters, we report final results by running inference on the test split.

\subsection{Training Details}
\label{ssec:training_details}
\begin{table*}[htb]
    \centering
    \ra{1.2}
    \begin{tabular}{@{} l l l l @{}}
    \toprule
     & Riverbed Segmentation & Chesapeake Land Cover & Sen1Floods11 \\
    \midrule
    Encoder & ResNet-50 & ResNet-50 & ResNet-50 \\
    Batch size & 64 & 64 & 64 \\
    Atrous rates & (3,6,9) & (3,6,9) & (3,6,9) \\
    Optimizer & Momentum(m=0.9) & Momentum(m=0.9) & Momentum(m=0.9) \\
    Decoder output stride & 4 & 4 & 4 \\
    Train crop size & 321x321 & 241x241 & 321x321 \\
    Output stride & 16 & 8 & 16 \\

    \addlinespace[1ex]
    Evaluation Metric & \specialcell[t]{Mean IoU of\\Riverbed class} & \specialcell[t]{Average of mean\\IoU of all classes} & \specialcell[t]{Mean IoU of\\Water class} \\

    \addlinespace[1ex]
    Number of train steps & \specialcell[t]{60k for Supervised\\120k for FixMatch} & \specialcell[t]{100k for Supervised\\120k for FixMatch} & \specialcell[t]{20k for Supervised\\120k for FixMatch} \\

    \addlinespace[1ex]
    Partial datasets & \specialcell[t]{1 sample of 100\% \\ 3 samples of 10\% \\ 5 samples of 1\%} & \specialcell[t]{1 sample of 100\% \\ 3 samples of 10\% \\ 5 samples of 1\%} & \specialcell[t]{1 sample of 100\% \\ 5 samples of 10\% } \\

    \addlinespace[1ex]
    \specialcell[t]{Learning rate\\weight decay} & \multicolumn{3}{c}{\specialcell[t]{Decided by a sweep over learning rate values [0.3, 0.1, 0.03, 0.01, 0.003,\\0.001] and weight decay values [0.001, 0.0001, 0.0003, 0.00001, 0.000001]\\for each model on each dataset.}}\\
    \bottomrule
    \end{tabular}
    \caption{Training details of Supervised and FixMatch learning methods for all three datasets.}
    \label{tab:training_details}
\end{table*}

\textbf{Supervised}:
We use the ResNet-50 model as our backbone encoder with the first 7x7 convolution layer replaced with two 3x3 convolution layers (as done in DeepLabv3+ \cite{Chen_2018_ECCV}).
We use batch normalization\cite{ioffe2015batch}, a batch size of $64$, and the Momentum optimizer with momentum set to $0.9$.
We also use an exponential moving average of model parameters which helps to stabilize the evaluation of the model throughout the training.
Atrous rates are set to $(3, 6, 9)$ for all experiments.
The skip connection between encoder and decoder is applied at the layer with down-sampling factor of $4$.
The learning rate is decayed with a polynomial schedule with power $0.9$ from its initial value to zero.
For each dataset, we set the output stride (the ratio of input image size and the spatial resolution of intermediate feature maps) and the input image size according to the dataset image size (see Table \ref{tab:training_details} for details).
Refer to Chen \emph{et al.} \cite{Chen_2018_ECCV} for more details on these hyperparameter values.

\textbf{SimCLR}:
The same ResNet-50 backbone that was used for the Supervised model is also used for SimCLR training to allow transferring weights.
For each dataset, SimCLR is trained on its corresponding unlabeled dataset with a setup similar to the SimCLR paper \cite{chen2020simple}.
We use the Layer-wise Adaptive Rate Scaling (LARS) optimizer with momentum $0.9$, weight decay of $0.0001$, a batch size of $4096$, an initial learning rate of $2.4$, and a cosine learning rate decay schedule.
These parameters were found to generally work well across all tasks, though it is possible that additional tuning for individual tasks may further increase performance.
For the riverbed segmentation task, each large unlabeled image of size 1024x1024 is split into tiles of size 256x256 and then a crop size of 224x224 was used to train for 160k steps.
For the Chesapeake land cover classification task, a crop size of 224x224 was used to train for 200k steps.
For the Sen1Floods11 flood extent mapping task, a crop size of 256x256 was used to train for 100k steps.
After SimCLR pre-training is done on the unlabeled data, the last checkpoint is used to initialize the model encoder for \supsimclr and \fmsimclr.

\textbf{FixMatch}:
For semi-supervised experiments, a hyperparameter sweep is run over the confidence threshold with values $[0.75, 0.8, 0.85, 0.9, 0.95]$.
A threshold of $0.9$ worked best for pseudo-label generation across all datasets.
The training batch consists of $32$ labeled and $32$ unlabeled images in all cases.
The total loss is calculated as the sum of supervised loss and semi-supervised loss, optimized using Momentum optimizer with momentum $0.9$ and a polynomial schedule with power $0.9$. Batch normalization parameters are updated only on the labeled and the strongly augmented images. Table \ref{tab:training_details} summarizes the other dataset-specific hyperparameter choices.

\textbf{Other details}:
To account for variability in optimal hyperparameter values across datasets and models, we perform a grid search to identify optimal learning rate values from \{3e-1, 1e-1, 3e-2, 1e-2, 3e-3, 1e-3\} and weight decay values from \{1e-3, 1e-4, 3e-4, 1e-5, 1e-6\} for each dataset and for each model except SimCLR pre-training.
The grid search is done on the $100\%$ sample of the domain-shifted partition and the same parameters are used for the $1\%$ and $10\%$ data experiments and all IID partition experiments.
The exact hyperparameters chosen per dataset are documented in the Appendix.
For each model, the checkpoint with the best evaluation metric on the validation set is chosen for testing.
The ResNet-50 backbone used in all models is fully convolutional \cite{Long_2015_CVPR}, which allows us to transfer weights from models trained on ImageNet and SimCLR using different image sizes, as well as evaluate on image sizes that are different from the training image sizes.
For datasets with non-RGB inputs, whenever we initialize the model from an ImageNet checkpoint, the first convolution layer weights for each non-RGB channel are initialized with the mean of RGB weights of the first convolution layer of the ImageNet checkpoint (averaged across channels).

\begin{table*}

    \centering
    \begin{tabular}{c c c c c c c c}
    \toprule

\multirow{2}{*}{Dataset} & \multirow{2}{*}{Model} & \multicolumn{3}{c}{Domain-shifted Partition} & \multicolumn{3}{c}{IID Partition}\\
\cmidrule(lr){3-5} \cmidrule(lr){6-8}
& & 1\% & 10\% & 100\% & 1\% & 10\% & 100\% \\
\midrule

\multirow{6}{*}{\specialcellmid[c]{Riverbed\\Segmentation\\(Riverbed mIoU)}}
& \suprandom & $38.93\pm3.0$ & $53.17\pm2.7$ & $55.45\pm0.8$ & $61.46\pm6.6$ & $77.67\pm1.5$ & $83.00\pm0.1$ \\
& \supimgnet & $47.68\pm5.1$ & $53.39\pm8.6$ & $60.87\pm2.3$ & $68.26\pm6.6$ & $80.47\pm0.9$ & $\mathbf{84.70\pm0.0}$ \\
& \supsimclr & $54.46\pm2.0$ & $63.24\pm1.1$ & $65.39\pm1.3$ & $71.14\pm7.1$ & $80.60\pm0.3$ & $84.57\pm0.1$ \\
& \fmrandom  & $46.17\pm2.7$ & $54.21\pm1.2$ & $58.72\pm0.5$ & $67.49\pm5.9$ & $77.78\pm1.1$ & $81.95\pm0.0$ \\
& \fmimgnet  & $54.76\pm2.6$ & $62.54\pm0.9$ & $64.27\pm0.7$ & $71.82\pm7.5$ & $\mathbf{81.88\pm0.3}$ & $84.31\pm0.1$ \\
& \fmsimclr  & $\mathbf{61.01\pm1.9}$ & $\mathbf{67.22\pm0.9}$ & $\mathbf{69.78\pm1.2}$ & $\mathbf{74.29\pm5.9}$ & $81.59\pm0.2$ & $84.39\pm0.0$ \\

\midrule

\multirow{6}{*}{\specialcellmid[c]{Chesapeake\\Land Cover\\(Avg. mIoU)}}
& \suprandom & $63.63\pm1.9$ & $72.06\pm1.0$ & $76.89\pm0.2$ & $79.06\pm2.3$ & $82.18\pm0.8$ & $83.41\pm0.2$ \\
& \supimgnet & $62.48\pm5.0$ & $73.65\pm1.0$ & $78.31\pm0.3$ & $80.60\pm1.1$ & $82.60\pm0.6$ & $83.65\pm0.1$ \\
& \supsimclr & $65.89\pm2.0$ & $74.05\pm0.8$ & $78.51\pm0.4$ & $80.03\pm1.5$ & $82.43\pm0.7$ & $\mathbf{84.25\pm0.3}$ \\
& \fmrandom &  $67.72\pm3.3$ & $72.23\pm1.6$ & $76.34\pm0.3$ & $80.71\pm2.0$ & $81.81\pm1.4$ & $82.54\pm0.2$ \\
& \fmimgnet &  $63.73\pm5.7$ & $75.38\pm1.7$ & $78.64\pm0.3$ & $80.86\pm1.3$ & $82.55\pm0.4$ & $83.80\pm0.1$ \\
& \fmsimclr &  $\mathbf{69.15\pm2.5}$ & $\mathbf{76.26\pm2.1}$ & $\mathbf{79.01\pm0.0}$ & $\mathbf{81.76\pm0.7}$ & $\mathbf{82.86\pm0.4}$ & $83.64\pm0.1$ \\

\midrule

\multirow{6}{*}{\specialcellmid[c]{Sen1Floods11\\(Water mIoU)}}
& \suprandom & & $60.11\pm10.3$ & $67.77\pm0.8$ & & $53.69\pm9.7$ & $63.51\pm1.2$ \\
& \supimgnet & & $61.91\pm7.6$ & $67.43\pm0.5$ & & $56.68\pm1.9$ & $64.79\pm0.5$ \\
& \supsimclr & & $64.29\pm3.9$ & $70.95\pm0.1$ & & $\mathbf{59.56\pm3.0}$ & $\mathbf{66.92\pm0.5}$ \\
& \fmrandom & & $66.05\pm3.5$ & $\mathbf{71.56\pm0.6}$ & & $56.24\pm5.2$ & $62.61\pm0.6$ \\
& \fmimgnet & & $\mathbf{66.46\pm3.6}$ & $71.07\pm0.3$ & & $59.17\pm3.4$ & $64.66\pm0.4$ \\
& \fmsimclr & & $64.08\pm6.3$ & $70.63\pm0.5$ & & $59.47\pm1.6$ & $61.35\pm2.4$ \\

    \bottomrule
    \end{tabular}

    \caption{Results on test set of the domain-shifted and IID partitions of all three datasets. The numbers show the aggregated mean and standard deviation of metrics.}
    \label{tab:all_results}
\end{table*}

\subsection{Results: Riverbed Segmentation}

\begin{figure*}[htb]
  \centering
  \subfigure[Domain-shifted partition\label{fig:results_river_hard_test}]{\includegraphics[scale=0.39]{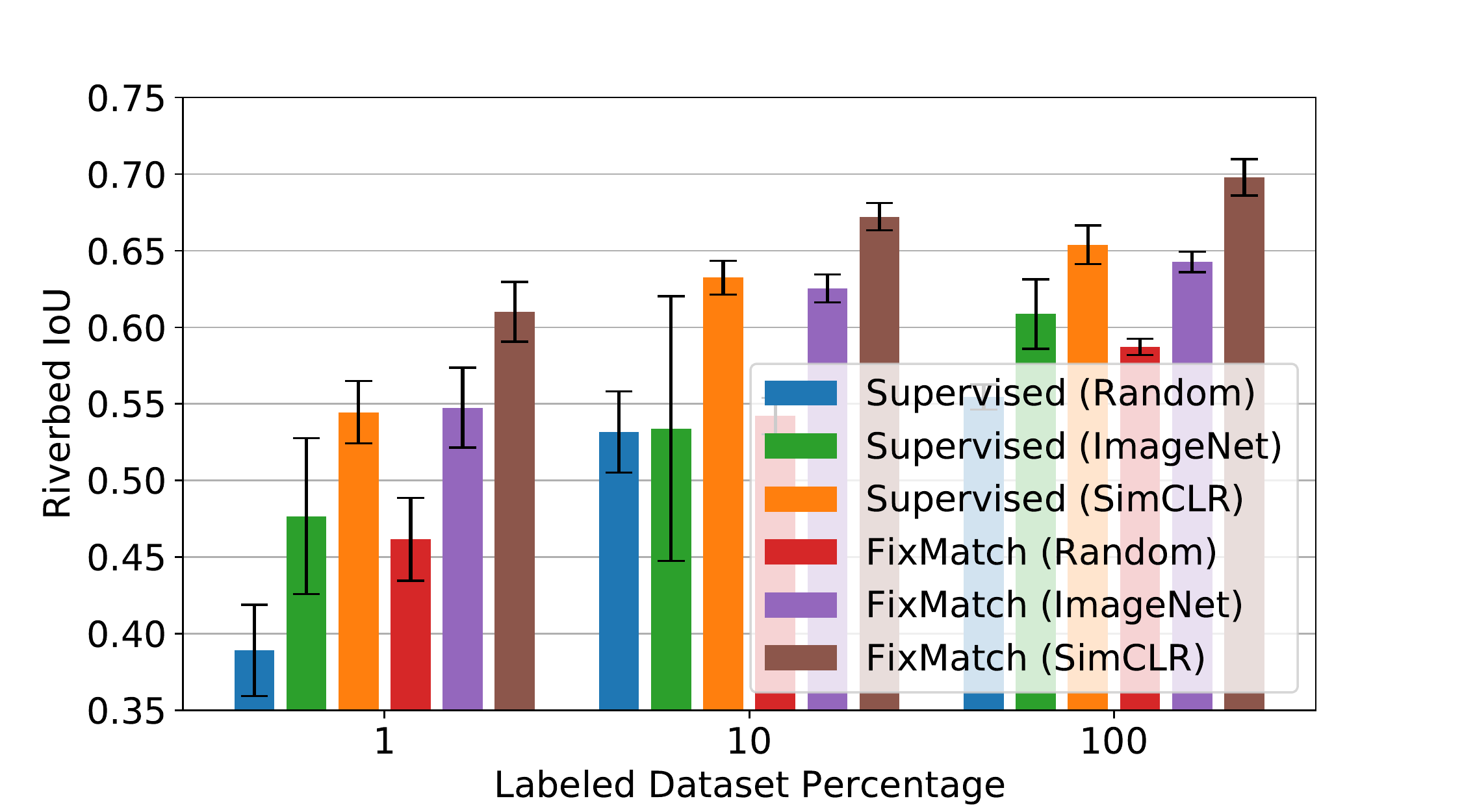}}\hfill
  \subfigure[IID partition\label{fig:results_river_iid_test}]{\includegraphics[scale=0.39]{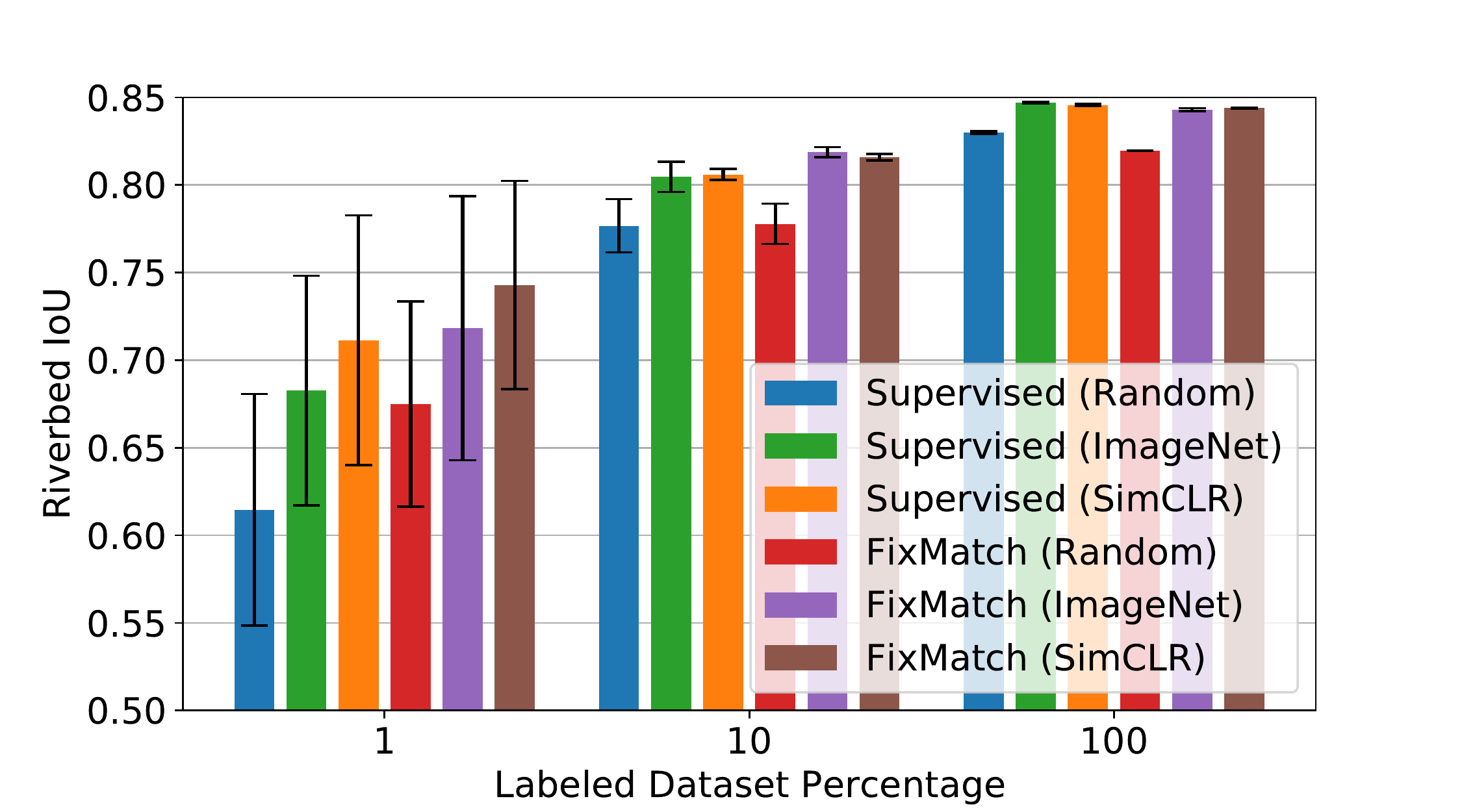}}
  \caption{Results on the test split of River segmentation dataset. Note that the y-axis range is different for the two dataset partitions as the performance range is significantly different between the two.}
\end{figure*}

Following the metric used in the Common Objects in Context (COCO) segmentation benchmark \cite{caesar2018cocostuff}, we use the pixel-wise Intersection over Union (IoU) of the riverbed class to validate our model performance on this dataset. Although this dataset has another labeled class (clouds), these annotations are ambiguous even for human annotators as some images contain hazy/semi-transparent clouds.
Furthermore, some of the validation and test splits have extremely small number of cloud pixels due to their geographical location, which made metrics on the cloud class unreliable.
Hence, we run our evaluation only on the riverbed class metric.

\subsubsection{Domain-shifted partition}

Figure \ref{fig:results_river_hard_test} shows riverbed IoU on the domain-shifted partition for \supsimclr, \fmrandom and \fmimgnet against the baselines \suprandom and \supimgnet. Because of in-domain pre-training, \supsimclr provides better model initialization and consistently outperforms \supimgnet (7\%, 10\% and 4.5\% absolute riverbed IoU improvement in 1\%, 10\% and 100\% labeled dataset respectively). For semi-supervised learning, \fmrandom provides an absolute riverbed IoU improvement over \suprandom of 7\% in 1\% dataset, going down to 3\% in 100\% dataset. \fmimgnet shows a consistent increase in the absolute riverbed IoU of 7\%, 9.1\% and 3.4\% over \supimgnet in 1\%, 10\% and 100\% of the labeled dataset respectively.

Our best model \fmsimclr, which combines SimCLR pre-training and FixMatch semi-supervised training, significantly outperforms all other models. \fmsimclr trained only on 1\% labeled dataset has a riverbed IoU of 61.0\%, which matches the commonly used \supimgnet baseline trained on full 100\% dataset (60.9\% riverbed IoU). This result suggests that combining self- and semi-supervised models can make deep learning extremely data efficient for general riverbed segmentation tasks.

\subsubsection{IID partition}

Figure \ref{fig:results_river_iid_test} compares the models on the IID partition of the riverbed dataset in a similar way as the domain-shifted partition. Initializing the model with ImageNet or SimCLR pre-trained weights always helps against random initialization for both supervised and semi-supervised models. When labeled data is available in enough quantity (i.e., 10\% and 100\%), model performance saturates and there isn’t much gain in using self- and semi-supervised techniques. However, for the 1\% labeled data sample, \supsimclr, \fmimgnet and \fmsimclr provide a 2.9\%, 3.6\% and 6\% absolute riverbed IoU improvement respectively over the \supimgnet baseline. It is interesting to note that on the 100\% dataset, \fmrandom performance degrades by 1.1\% on absolute riverbed IoU from the \suprandom baseline. Our hypothesis is that the FixMatch model might be losing some accuracy on the labeled training data, which is sampled from the same region as the test data, while learning to generalize better on other regions of the unlabeled dataset.

\subsection{Results: Chesapeake Land Cover}

\begin{figure*}[htb]
  \centering
  \subfigure[Domain-shifted partition\label{fig:results_chesa_hard_test}]{\includegraphics[scale=0.39]{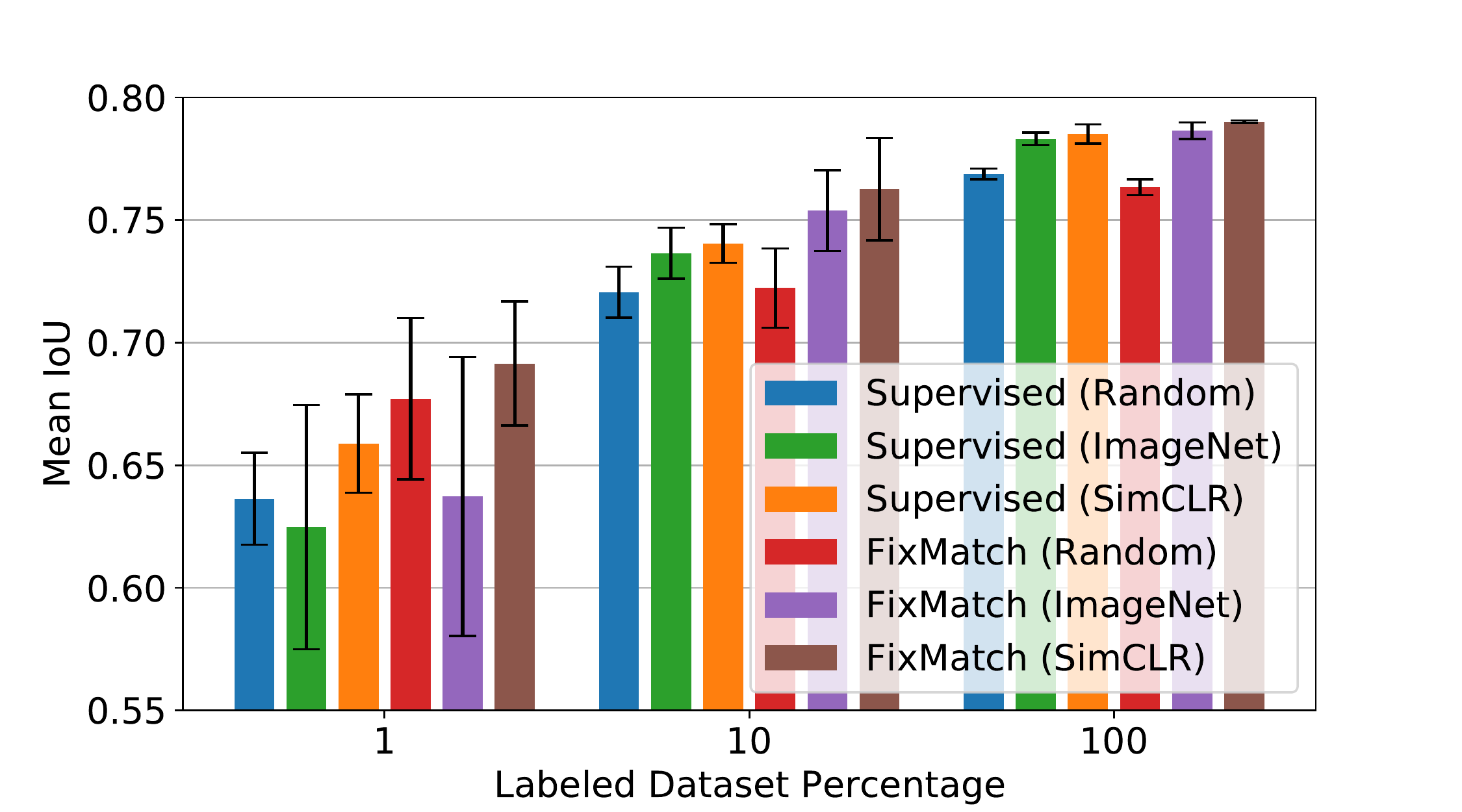}}\hfill
  \subfigure[IID partition\label{fig:results_chesa_iid_test}]{\includegraphics[scale=0.39]{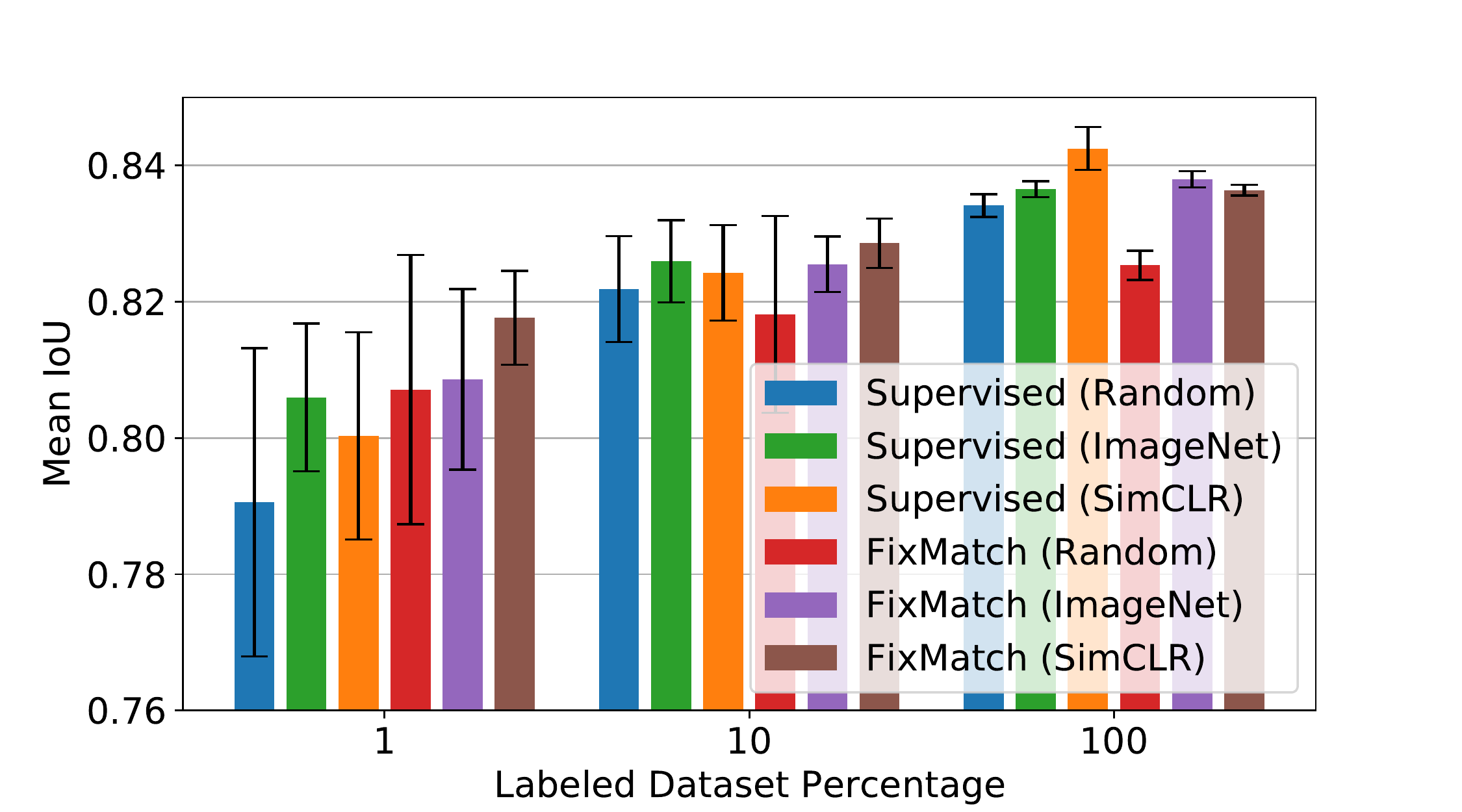}}
  \caption{Results on the test split of Chesapeake dataset. Note that the y-axis range is different for the two dataset partitions as the performance range is significantly different between the two.}
\end{figure*}

We use mean IoU (average of the pixel-wise IoU of all four classes) to validate our model performance on this dataset as done by \cite{robinson2019large}. We used different unlabeled datasets for the SimCLR and FixMatch models for this task (see Figure \ref{fig:chesapeake_unlabeled}). Unlabeled data from the eastern half of the US resulted in better performance for the FixMatch models compared to using data sampled across the conterminous US. Our hypothesis for this behavior is that semi-supervised learning benefits more from unlabeled data that is closer in distribution to the eventual test data.

\subsubsection{Domain-Shifted partition}

Figure \ref{fig:results_chesa_hard_test} shows the mean IoU on the domain-shifted partition of this dataset for all models. \supsimclr performs slightly better than \supimgnet baseline for the 10\% and the 100\% datasets. However, the benefit of SimCLR pre-training is more pronounced for the 1\% labeled dataset where it outperforms \supimgnet by an absolute margin of 3.4\% in mean IoU and \suprandom by 2.3\% in mean IoU (9.1\% and 6.2\% relative reduction in error rate).

\fmimgnet provides an absolute improvement of by 1.25\% and 1.73\% in the 1\% and 10\% datasets, respectively, over \supimgnet. For the 100\% dataset however, there are no notable gains in using \fmimgnet over the baseline \supimgnet. \fmrandom performs better than \suprandom by 4.1\% for the 1\% dataset but is similar in performances for the 10\% and 100\% datasets.

The combined \fmsimclr model outperforms all other models and results in absolute mean IoU improvements of 6.7\%, 2.6\%, 0.7\% over \supimgnet for the 1\%, 10\% and 100\% datasets, respectively.

\subsubsection{IID partition}

Figure \ref{fig:results_chesa_iid_test} compares all models on the IID partition of this dataset. \supsimclr outperforms \supimgnet in 100\% by a small margin of 0.6\% absolute mean IoU, but lags behind the \supimgnet baseline by 0.6\% for the 1\% dataset. \fmimgnet and \fmsimclr models, on the other hand, match the \supimgnet baseline for the 100\% dataset and outperform these models for the 1\% dataset by a small margin (0.3\% and 1.2\% absolute mean IoU improvement, respectively).

Overall from Figure \ref{fig:results_chesa_iid_test}, we do not see much gains in using unlabeled data techniques over the supervised baselines in this partition. The change in absolute performance of \supimgnet between 1\% and 100\% of labeled data is also fairly small (from 80.6\% to 83.65\%). Our hypothesis is that in the IID setting, this task is easier than the domain-shifted partition, and there are sufficient labels to achieve high accuracies even for the 1\% sample size (which corresponds to 500 labeled images). In addition, in our qualitative review, we found that the ground truth labels on Chesapeake have some errors, and it's possible that at 100\% sample size, we might be at the limit of ground truth accuracy and cannot see further model improvements.

\subsection{Results: Sen1Floods11}

\begin{figure*}[htb]
  \centering
  \subfigure[Domain-shifted partition\label{fig:results_sen_hard_test}]{\includegraphics[scale=0.39]{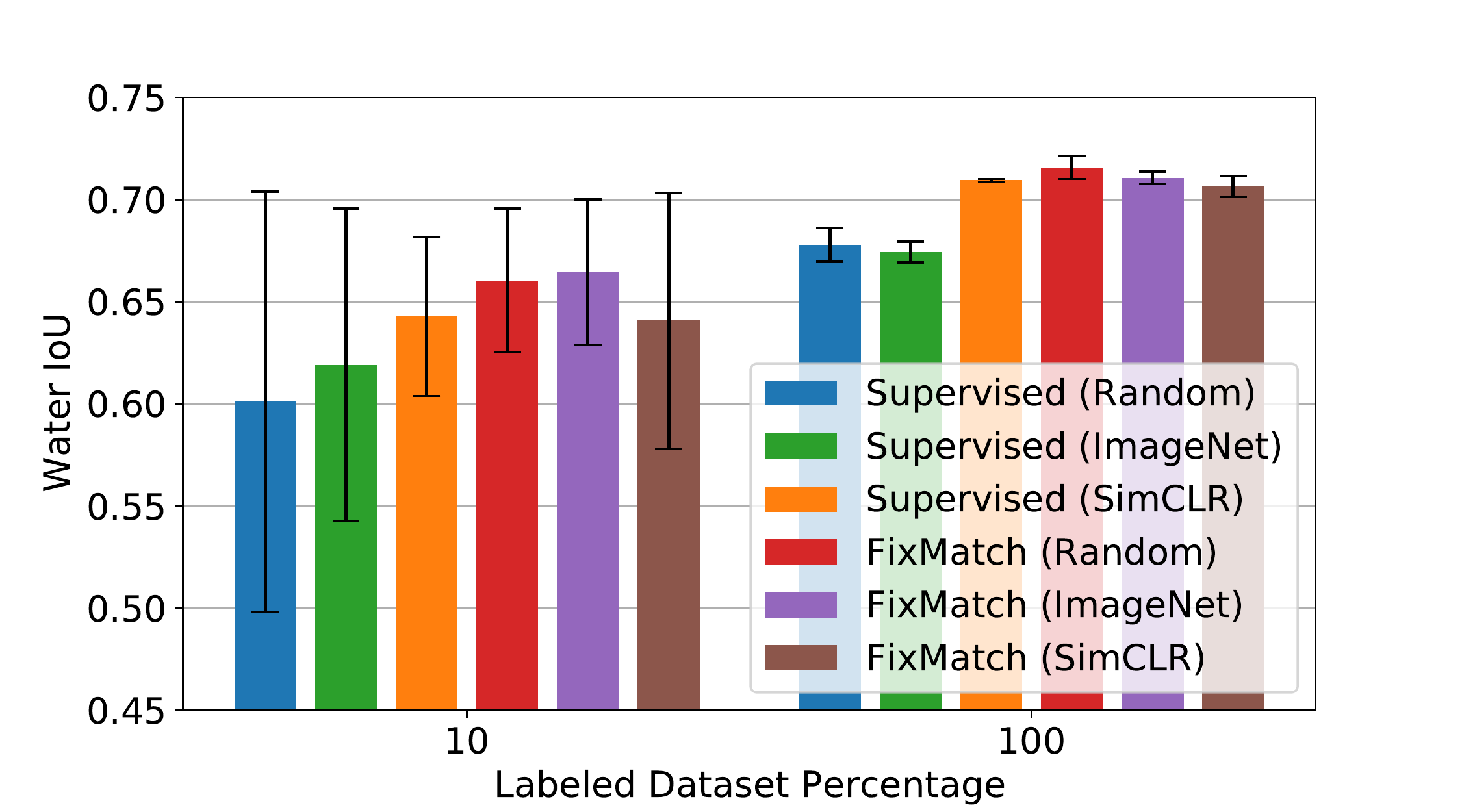}}\hfill
  \subfigure[IID partition\label{fig:results_sen_iid_test}]{\includegraphics[scale=0.39]{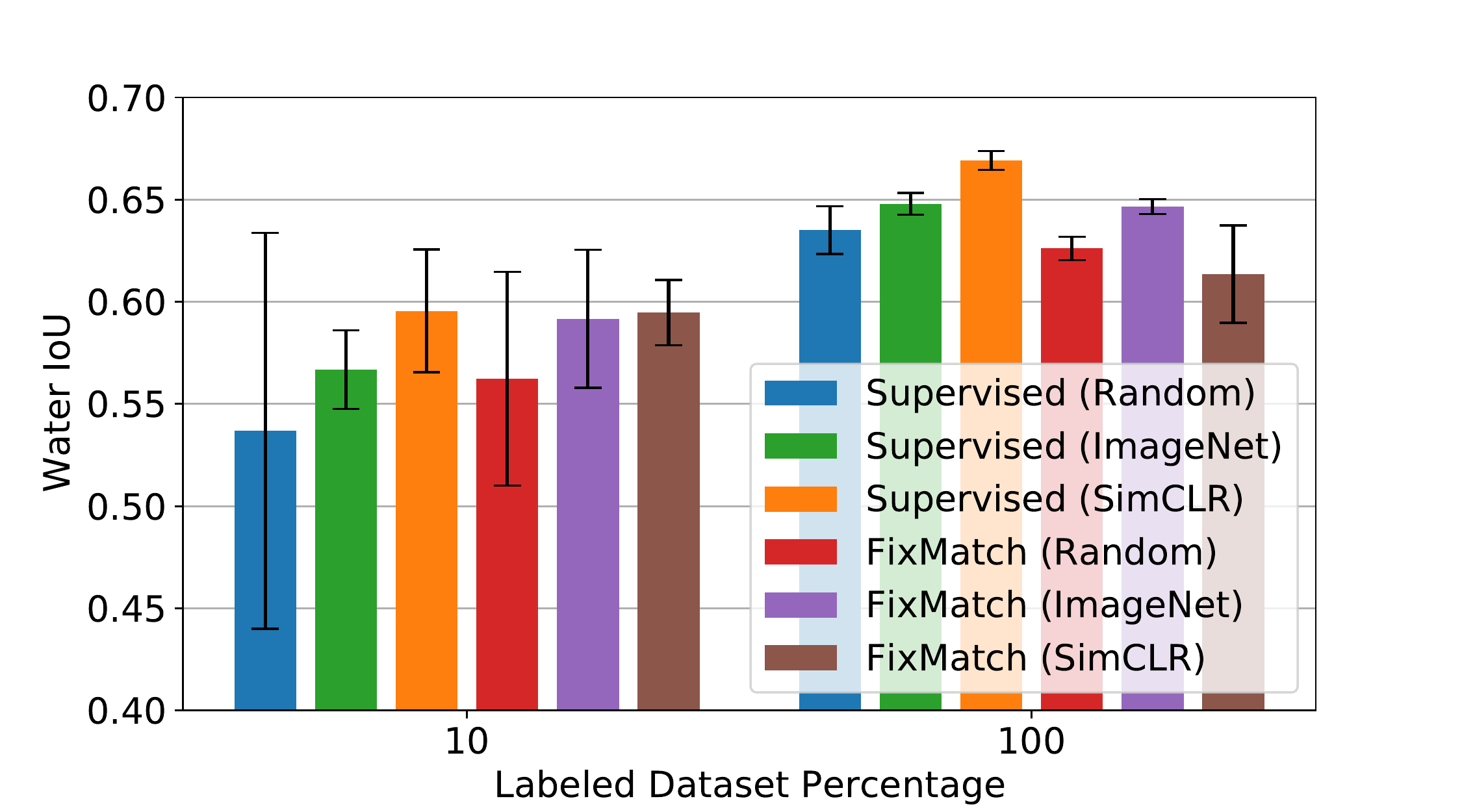}}
  \caption{Results on the test split of Sen1Floods11 dataset. Note that the y-axis range is different for the two dataset partitions as the performance range is slightly different between the two.}
\end{figure*}

The authors of the Sen1Floods11 dataset used the mean of image-wise IoU of the water class as their evaluation metric \cite{Bonafilia_2020_CVPR_Workshops}.
To keep our metric consistent across our tasks, we used the pixel-wise mean IoU of the water class to validate our model performance on this dataset. Unlike the image-wise IoU, this metric is not sensitive to the choice of the tile size and proportion of water pixels in a particular image.

\subsubsection{Domain-Shifted partition}

Figure \ref{fig:results_sen_hard_test} compares the performance of all models (using IoU of water class) on the domain-shifted partition of the Sen1Floods11 dataset. \supsimclr provided modest gains of 2.4\% and 3.5\% absolute IoU over the \supimgnet baseline for the 10\% and 100\% datasets, respectively.

\fmrandom outperforms \suprandom by 6.0\% and 3.8\% absolute IoU on the 10\% and 100\% datasets, respectively. With \fmimgnet, we see a gain of 4.55\% and 3.6\% absolute IoU over \supimgnet on the 10\% and 100\% datasets, respectively. \fmsimclr shows an improvement of 2.2\%, 3.2\% over \supimgnet for 10\% and 100\% datasets, respectively, matching \supsimclr in each case, but not providing any additional gains.

\subsubsection{IID partition}

Figure \ref{fig:results_sen_iid_test} compares the performance of all models on the IID partition of Sen1Floods11 dataset. Similar to the domain shifted partition, \supsimclr provided modest gains of 2.9\% and 2.1\% absolute IoU over the \supimgnet baseline for the 10\% and 100\% datasets, respectively.

Semi-supervised learning shows mixed results on this partition. \fmrandom outperforms \suprandom by 2.5\% on the 10\% dataset but shows a small degradation of 0.9\% for the 100\% dataset. The other variants of FixMatch (i.e. \fmimgnet and \fmsimclr) also show mixed performance. \fmimgnet improves performance over \supimgnet by 2.5\% for the 10\% dataset and matches it for the 100\% dataset. \fmsimclr matches \supsimclr in performance for the 10\% dataset but decreases performance by 5.4\% for the 100\% dataset.


%% file: sections/conclusion.tex
We evaluated the efficacy of using self- and semi-supervised methods on three different remote sensing segmentation tasks given two real world settings: (i) limited amounts of labeled data and (ii) geographical domain shifts. 
In scenarios where the performance of the purely supervised model saturates, signifying that there is enough labeled data for good generalization  (e.g. the IID partition with 100\% of labels for the Chesapeake land cover segmentation task), we found that self- and semi-supervised approaches do not provide meaningful benefits and FixMatch can sometimes result in slight decreases in performance compared to supervised baselines. However, one consistent trend observed across all tasks was that in settings with a small number of labels, using both SimCLR and FixMatch techniques separately and in combination improved model performance by a large margin. The gains were even more pronounced when the model was applied to new geographic domains at test time, indicating that these techniques are effective in leveraging unlabeled data to improve model generalization. Such improved capacity for generalization is especially useful in real-world applications where it may be desirable to apply models trained on labeled datasets from one location to new locations, i.e., generating updated flood maps as new events occur or using detailed land cover annotations to produce maps for new regions outside the initial labeling domain. Therefore, in real world deployments where some amount of geographical domain shift is expected, using the self- and semi-supervised techniques evaluated in this paper is expected to improve model generalization significantly.
\section*{Acknowledgments}

We would like to thank Vishal Batchu for insightful discussions, Umangi Jain for reviewing our experiments, Aparna Taneja for the help in generating the Sentinel-1 unlabeled data, and Oliver Guinan, Noel Gorelick, Christopher Brown, Wanda Czerwinski, and John Platt for reading drafts of this paper and providing feedback.

%% file: sections/appendix.tex
\appendices  
\label{sec:appendix}

\section{Hyperparameters}

Learning rate and weight decay significantly affect the model performance, especially for domain-shifted partition.
Hence, we perform a hyperparameter grid search and arrive at the following parameters for each model on each dataset.
All these hyper-parameters are selected using validation metric score of the 100\% dataset experiment of domain-shifted partition.

\subsection{Hyperparameters for Riverbed Segmentation Dataset}
\begin{center}
    \begin{tabular}{ c c c } 
        Model & Learning rate & Weight decay \\
        \midrule
        \suprandom & 0.01  & 0.001 \\
        \supimgnet & 0.03 & 0.0001 \\
        \supsimclr & 0.003 & 0.0001 \\
        \fmrandom & 0.03 & 0.0001 \\
        \fmimgnet & 0.01 & 0.00001  \\
        \fmsimclr & 0.003 & 0.0001 \\
    \end{tabular}
\end{center}

\subsection{Hyperparameters for Chesapeake Land Cover Dataset}
\begin{center}
    \begin{tabular}{ c c c } 
        Model & Learning rate & Weight decay \\
        \midrule
        \suprandom & 0.03  & 0.0001 \\
        \supimgnet & 0.01 & 0.0003 \\
        \supsimclr & 0.03 & 0.0003 \\
        \fmrandom & 0.03 & 0.0003 \\
        \fmimgnet & 0.03 & 0.0001  \\
        \fmsimclr & 0.01 & 0.0001 \\
    \end{tabular}
\end{center}

\subsection{Hyperparameters for Sen1Floods11 Dataset}
\begin{center}
    \begin{tabular}{ c c c } 
        Model & Learning rate & Weight decay \\
        \midrule
        \suprandom & 0.01  & 0.0001 \\
        \supimgnet & 0.001 & 0.000001 \\
        \supsimclr & 0.01 & 0.0001 \\
        \fmrandom & 0.03 & 0.0001 \\
        \fmimgnet & 0.03 & 0.0001  \\
        \fmsimclr & 0.1 & 0.0001 \\
    \end{tabular}
\end{center}

\section{Augmentations}

\subsection{Supervised Training Augmentations}
\label{appendix:deeplab_aug}
We use the following augmentation policy for supervised training of the DeepLabv3+ models: \suprandom, \supimgnet and \supsimclr.
We also use this same policy for weak augmentation in all FixMatch models.
\begin{itemize}
    \item Random distorted crop with 0.5 distortion.
    \item Random horizontal flip
    \item Random vertical flip
    \item Random rotation by a multiple of 90 degrees
    \item Color jitter with 0.5 probability:
    \begin{itemize}
        \item \texttt{ColorJitterRGB} with strength 0.4 for RGB images
        \item \texttt{ColorJitterGeneral} with strength 0.4 for non-RGB images
    \end{itemize}
\end{itemize}
Refer to Listing-\ref{lst:aug_code} below for the definition of \texttt{ColorJitterRGB} and \texttt{ColorJitterGeneral}.

\subsection{Ablation on Supervised Learning Augmentation}

Our augmentation policy for \supimgnet has 2 hyperparameters - distortion of crop and strength of color jitter.
On an initial ablation done on Sen1Floods11 dataset, the crop distortion of 0.5 and color jitter strength of 0.4 worked the best.
With this defined augmentation policy, we carried out a formal ablation on the effect of augmentation on \supimgnet baseline model performance.
For domain-shifted partition of Riverbed segmentation dataset and Sen1Floods11 dataset, we trained two more models: (1) \supimgnet with color jitter removed (i.e. no appearance augmentation) and (2) \supimgnet with all augmentation removed (no geometric and appearance augmentation).
Hyperparameters were chosen again for these new models with the same hyperparameter sweeps as discussed in Section \ref{ssec:training_details}.
The results on test splits are shown in the Figure \ref{fig:augmentation_ablation} below.
The figures clearly demonstrate that the chosen set of augmentations significantly improve the overall performance of \supimgnet across datasets and percentage splits, making it a strong baseline.

\begin{figure}[htb]
    \centering
    \includegraphics[width=\linewidth]{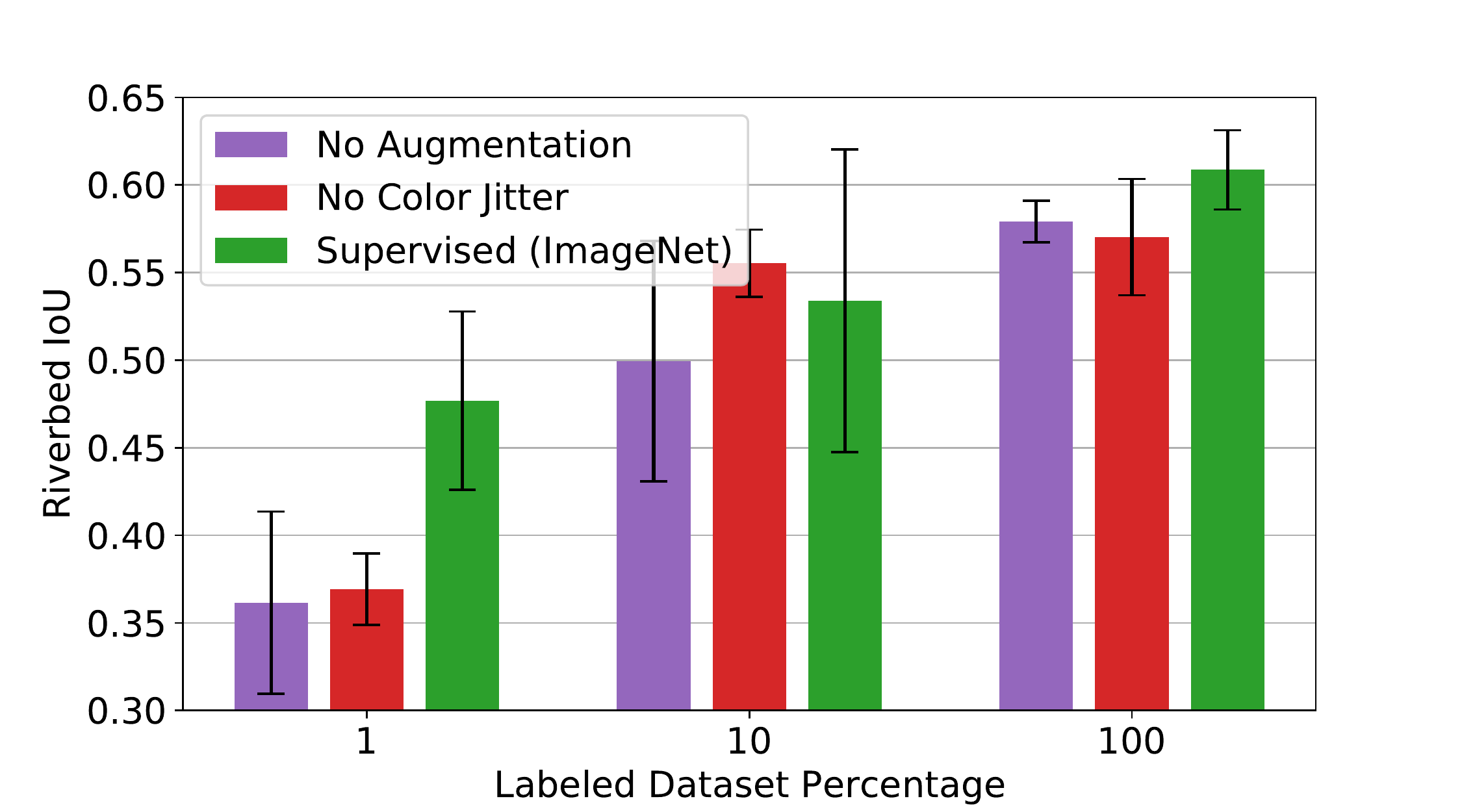}
    \includegraphics[width=\linewidth]{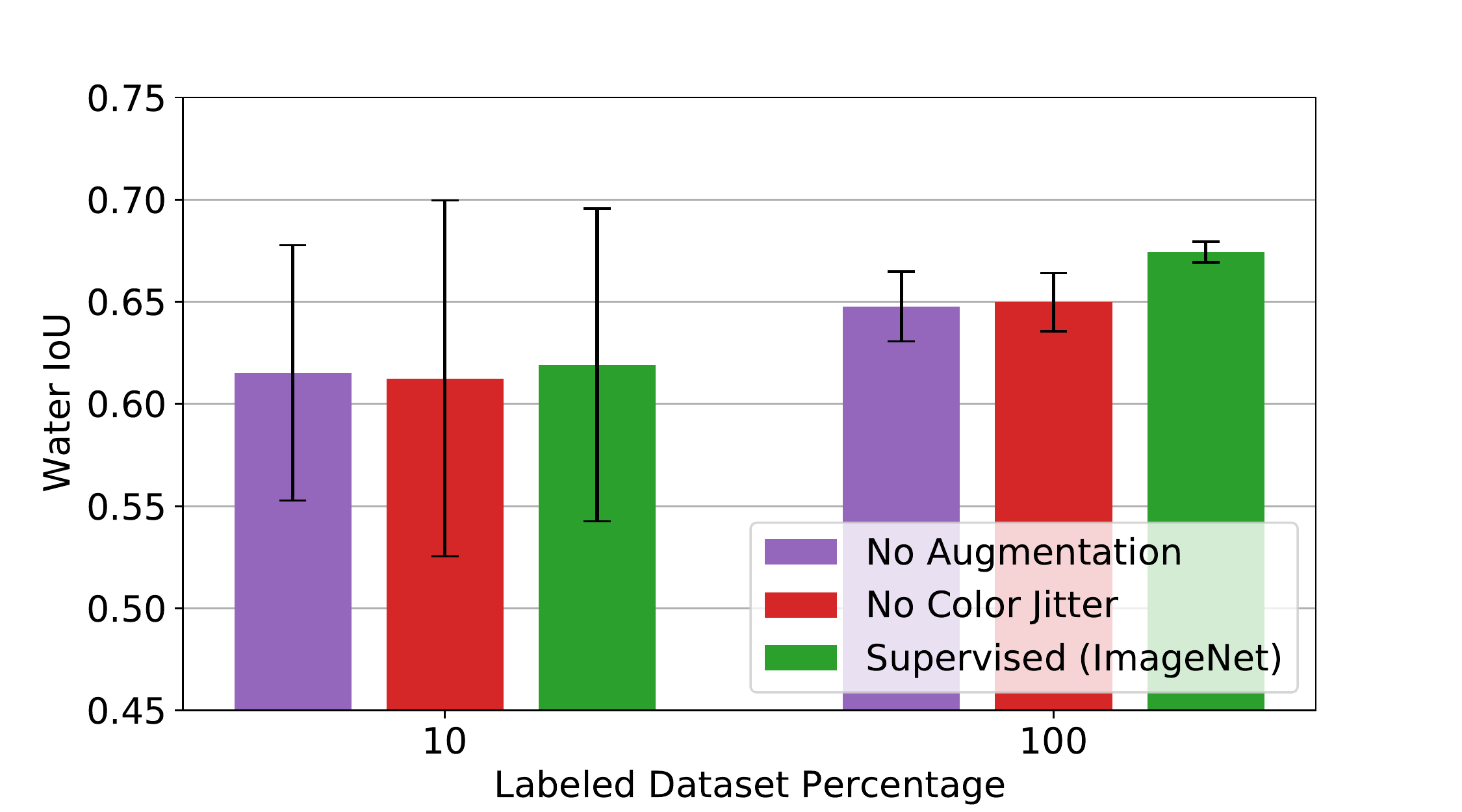}
    \caption{Augmentations ablation on the domain-shifted partition of Riverbed segmentation(top) and Sen1Floods11(bottom) dataset.}
    \label{fig:augmentation_ablation}
\end{figure}

\subsection{SimCLR Augmentations}

For RGB images, we use the same augmentations proposed by SimCLR\cite{chen2020simple}.
For non-RGB images, we replace RGB color jitter with general color jitter.
For color dropping, we use the mean of all channels for Chesapeake Land Cover dataset (RGB+IR images).
But the notion of color dropping is not defined for Sentinel-1 images.
Hence for Sen1Floods11, we don’t use color dropping and instead adjust color jitter probability to $0.85$. Below is the full list of SimCLR augmentations used:
\begin{itemize}
    \item Distorted bounding box crop
    \item Random horizontal flip
    \item Random vertical flip
    \item Color jitter with 0.8 probability (probability 0.85 for Sen1Floods11)
    \begin{itemize}
        \item \texttt{ColorJitterRGB} with strength 0.8 for RGB images
        \item \texttt{ColorJitterGeneral} with strength 0.8 for non-RGB images
    \end{itemize}
    \item Color dropping with 0.2 probability
    \begin{itemize}
        \item Riverbed segmentation: tf.image.rgb\_to\_grayscale for RGB images
        \item Chesapeake Land Cover: Taking mean of all channels
        \item Sen1Floods11: No-operation
    \end{itemize}
    \item Random Gaussian blur
\end{itemize}
Refer to Listing-\ref{lst:aug_code} below for the definition of \texttt{ColorJitterRGB} and \texttt{ColorJitterGeneral}.

\subsection{FixMatch Augmentations}

Augmentations are used for all 3 version of images during the training pass: labeled, weakly-augmented and strongly-augmented. For the labeled images and weakly augmented images, the augmentation policy for supervised DeepLabv3+ models (defined in Appendix \ref{appendix:deeplab_aug}) is used.
For strong augmentations, two simple augmentation functions are applied successively, followed by a random cutout to generate a strong augmentation. For the cutout augmentation, while FixMatch\cite{sohn2020fixmatch} only removes a single large rectangle, we remove multiple smaller rectangles.
Table \ref{tab:strong_aug} lists the compositions of these augmentations.
For each image one of the composition is randomly selected and the individual functions are applied successively.

\begin{table*}
    \centering
    \begin{tabular}{r c l}
        \toprule
        Function & Parameter & Description \\
        \midrule
        AutoContrast &   & Maximizes the image contrast by setting the darkest (lightest) pixel to black (white). \\
        Brightness & $S$ & Change the brightness of all channels by a random factor in $[1-S, 1+S]$. \\
        Color & $C$ & Controls the contrast of the image. A $C=0$ returns a gray image, $C=1$ returns the original image. \\
        Contrast & $S$ & Change the contrast of all channels by a random factor in $[1-S, 1+S]$. \\
        Equalize &   & Equalizes the image histogram. \\
        Hue & $S$ & Change the hue of RGB image by a random factor in $[1-S, 1+S]$. \\
        Invert &   & Adjusts each pixel value $p$ to $(255-p)$ \\
        Posterize & $B$ & Reduces each pixel to $B$ bits. \\
        Rotate & $\theta$ & Rotates the image by $\theta$ degrees. \\
        Saturation & $S$ & Change the saturation of RGB image by a random factor in $[1-S, 1+S]$. \\
        ShearX & $R$ & Shears the image along the horizontal axis with rate $R$. \\
        ShearY & $R$ & Shears the image along the vertical axis with rate $R$. \\
        Solarize & $T$ & Inverts all pixels above a threshold value of $T$. \\
        SolarizeAdd & $A,T$ & Add $A$ to each pixel then invert each pixel with value above the threshold $T$.  \\
        TranslateX & $\lambda$ & Translates the image horizontally by $(\lambda \times image\_width)$ pixels. \\
        TranslateY & $\lambda$ & Translates the image vertically by $(\lambda \times image\_height)$ pixels. \\
        \bottomrule
    \end{tabular}
    \caption{Functions used in data augmentation policies for training the models.}
    \label{tab:aug_description}
\end{table*}

\begin{table}
    \centering
    \begin{tabular}{l l}
        \toprule
        Augmentation 1 & Augmentation 2 \\
        \midrule
        Equalize(0.8, .1) & ShearY(0.8, 0.4) \\
        Color(0.4, .9) & Equalize(0.6, 0.3) \\
        Color(0.4, .1) & Rotate(0.6, 0.8) \\
        Solarize(0.8, .3) & Equalize(0.4, 0.7) \\
        Solarize(0.4, .2) & Solarize(0.6, 0.2) \\
        Color(0.2, .0) & Equalize(0.8, 0.8) \\
        Equalize(0.4, .8) & SolarizeAdd(0.8, 0.3) \\
        ShearX(0.2, .9) & Rotate(0.6, 0.8) \\
        Color(0.6, .1) & Equalize(1.0, 0.2) \\
        Invert(0.4, .9) & Rotate(0.6, 0.0) \\
        Equalize(1.0, .9) & ShearY(0.6, 0.3) \\
        Color(0.4, .7) & Equalize(0.6, 0.0) \\
        Posterize(0.4, .6) & Autocontrast(0.4, 0.7) \\
        Solarize(0.6, .8) & Color(0.6, 0.9) \\
        Solarize(0.2, .4) & Rotate(0.8, 0.9) \\
        Rotate(1.0, .7) & TranslateY(0.8, 0.9) \\
        ShearX(0.0, .0) & Solarize(0.8, 0.4) \\
        ShearY(0.8, .0) & Color(0.6, 0.4) \\
        Color(1.0, .0) & Rotate(0.6, 0.2) \\
        Equalize(0.8, .4) & Equalize(0.0, 0.8) \\
        Equalize(1.0, .4) & Autocontrast(0.6, 0.2) \\
        ShearY(0.4, .7) & SolarizeAdd(0.6, 0.7) \\
        Posterize(0.8, .2) & Solarize(0.6, 1.0) \\
        Solarize(0.6, .8) & Equalize(0.6, 0.1) \\
        Color(0.8, .6) & Rotate(0.4, 0.5) \\
        \bottomrule
    \end{tabular}
    \caption{Augmentation strategy proposed in AutoAugment \cite{cubuk2019autoaugment} used as strong augmentation for FixMatch training. Each augmentation is specified as $function(probability,strength)$. For non-RGB inputs like Sen1Floods11 we replace the  \texttt{Color} augmentation with \texttt{ColorJitterGeneral}.}
    \label{tab:strong_aug}
\end{table}

\subsection{Pseudo Code}

We follow AutoAugment\cite{cubuk2019autoaugment} and SimCLR\cite{chen2020simple} for the implementation of most augmentation functions.
Please refer to the papers and their public implementations for specific details.
Listing-\ref{lst:aug_code} contains pseudo code for remaining augmentation functions that we use. 

\definecolor{codegreen}{rgb}{0,0.6,0}
\definecolor{codegray}{rgb}{0.5,0.5,0.5}
\definecolor{codepurple}{rgb}{0.58,0,0.82}
\definecolor{backcolour}{rgb}{0.95,0.95,0.92}
\lstdefinestyle{mystyle}{
    backgroundcolor=\color{backcolour},   
    commentstyle=\color{codegreen},
    keywordstyle=\color{magenta},
    numberstyle=\tiny\color{codegray},
    stringstyle=\color{codepurple},
    basicstyle=\ttfamily\footnotesize,
    breakatwhitespace=false,         
    breaklines=true,                 
    captionpos=t,                    
    keepspaces=true,                 
    numbers=left,                    
    numbersep=5pt,                  
    showspaces=false,                
    showstringspaces=false,
    showtabs=false,                  
    tabsize=2
}
\lstset{style=mystyle}
\begin{lstlisting}[
    basicstyle=\ttfamily\footnotesize,
    language=Python,
    caption={Pseudo-code for augmentations using Tensorflow},
    numbers=none,
    label={lst:aug_code}
]

def BrightnessPerChannel(image, strength):
  channels_list = split_channels(image)
  aug_channels_list = [
    Brightness(channel, strength) for channel in channels_list]
  return stack_channels(aug_channels_list)

def ContrastPerChannel(image, strength):
  channels_list = split_channels(image)
  aug_channels_list = [
    Contrast(channel, strength) for channel in channels_list]
  return stack_channels(aug_channels_list)

def ColorJitterRGB(image, strength):
  # We also randomize the order of these
  # augmentations each time.
  image = Brightness(image, strength)
  image = Contrast(image, strength)
  image = Hue(image, strength/4)
  image = Saturation(image, strength)
  return image

def ColorJitterGeneral(image, strength):
  # We also randomize the order of these
  # augmentations each time.
  image = Brightness(image, strength)
  image = Contrast(image, strength)
  image = BrightnessPerChannel(image, strength/2)
  image = ContrastPerChannel(image, strength/2)
  return image

\end{lstlisting}

\section{Validation split Plots}
\begin{figure*}
    \begin{subfigure}
        \centering
        \includegraphics[width=0.45\linewidth]{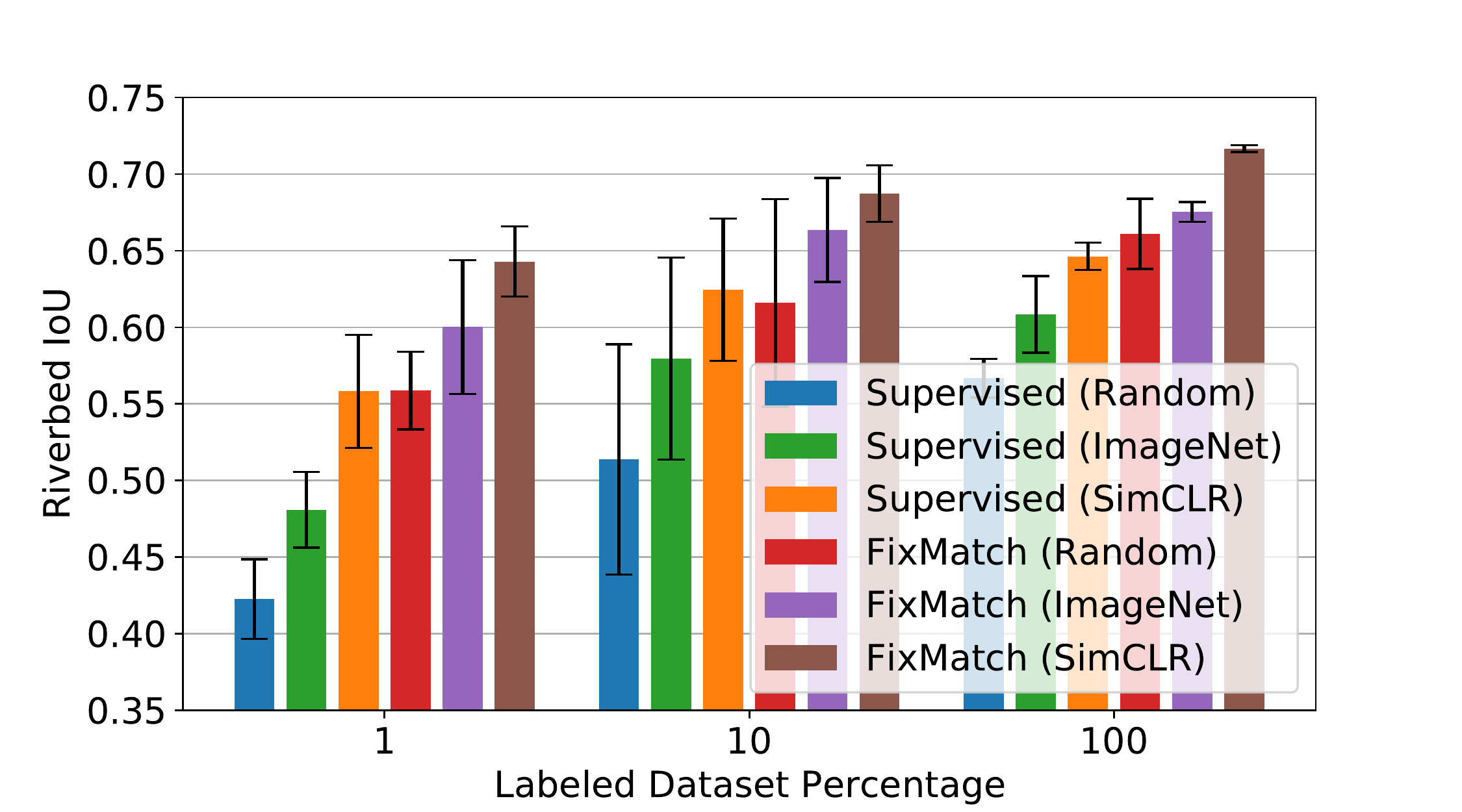}
        \includegraphics[width=0.45\linewidth]{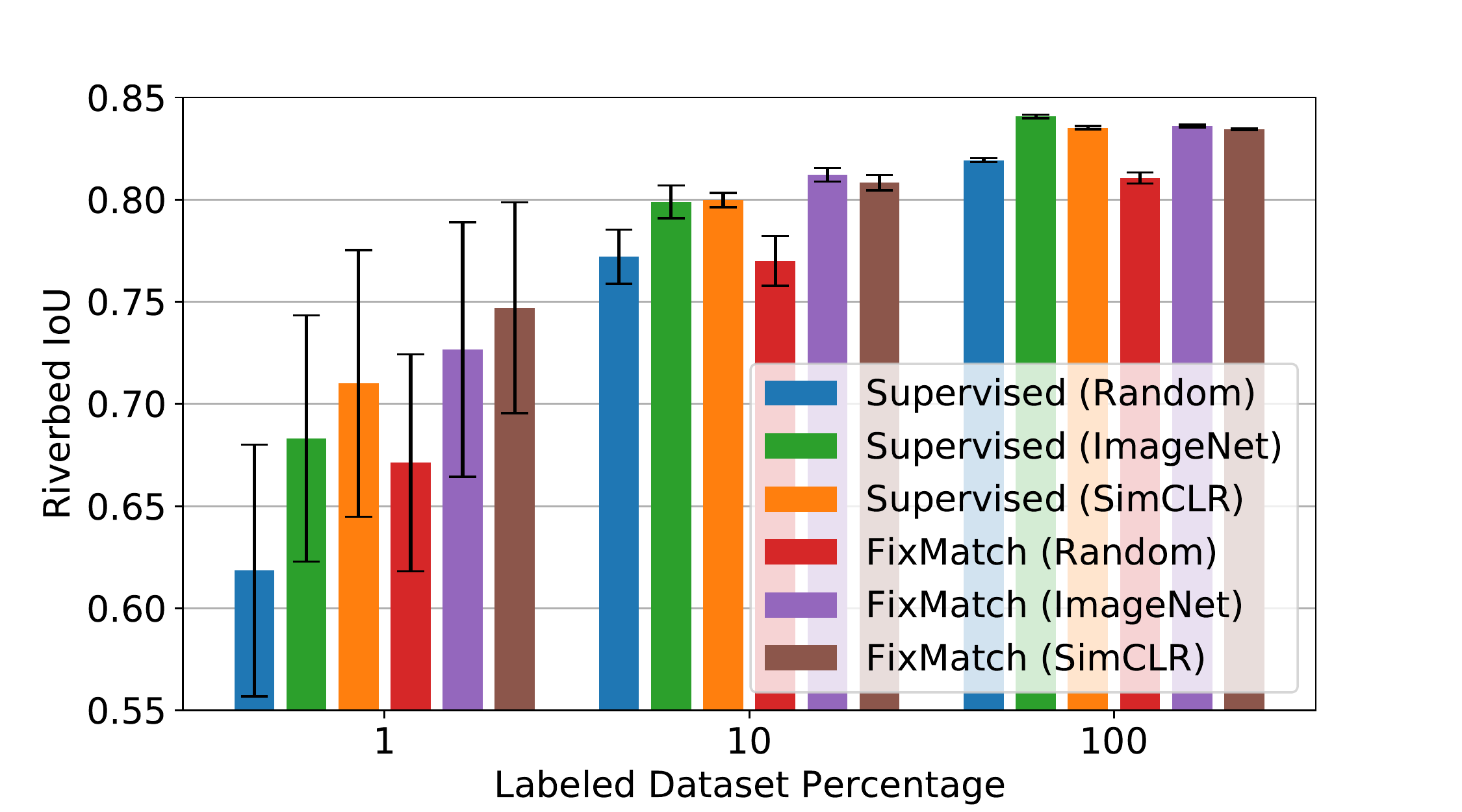}
        \caption{Results on the validation split of River segmentation domain-shifted partition (left) and IID partition (right).}
        \label{fig:results_river_val}
    \end{subfigure}
    \begin{subfigure}
        \centering
        \includegraphics[width=0.45\linewidth]{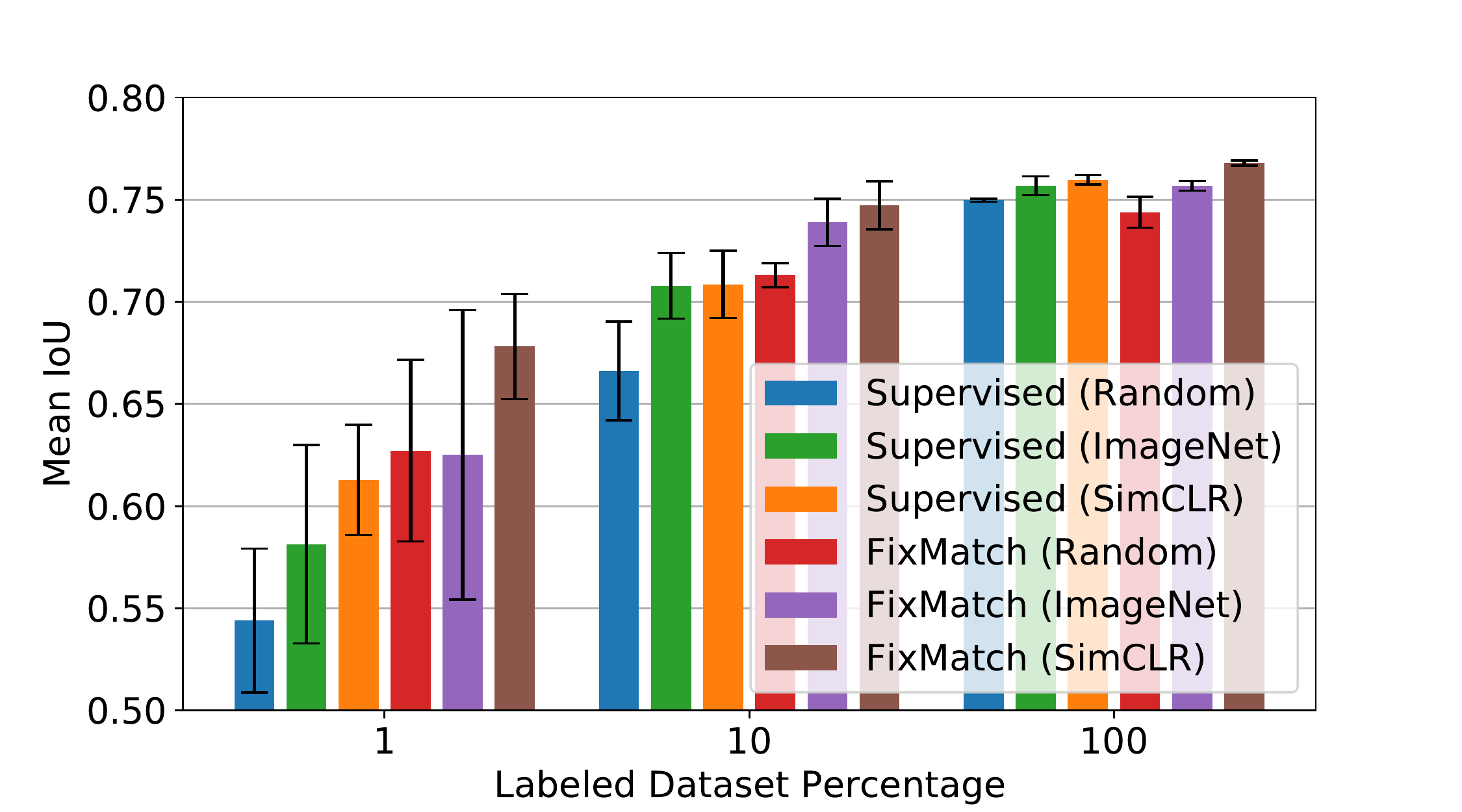}
        \includegraphics[width=0.45\linewidth]{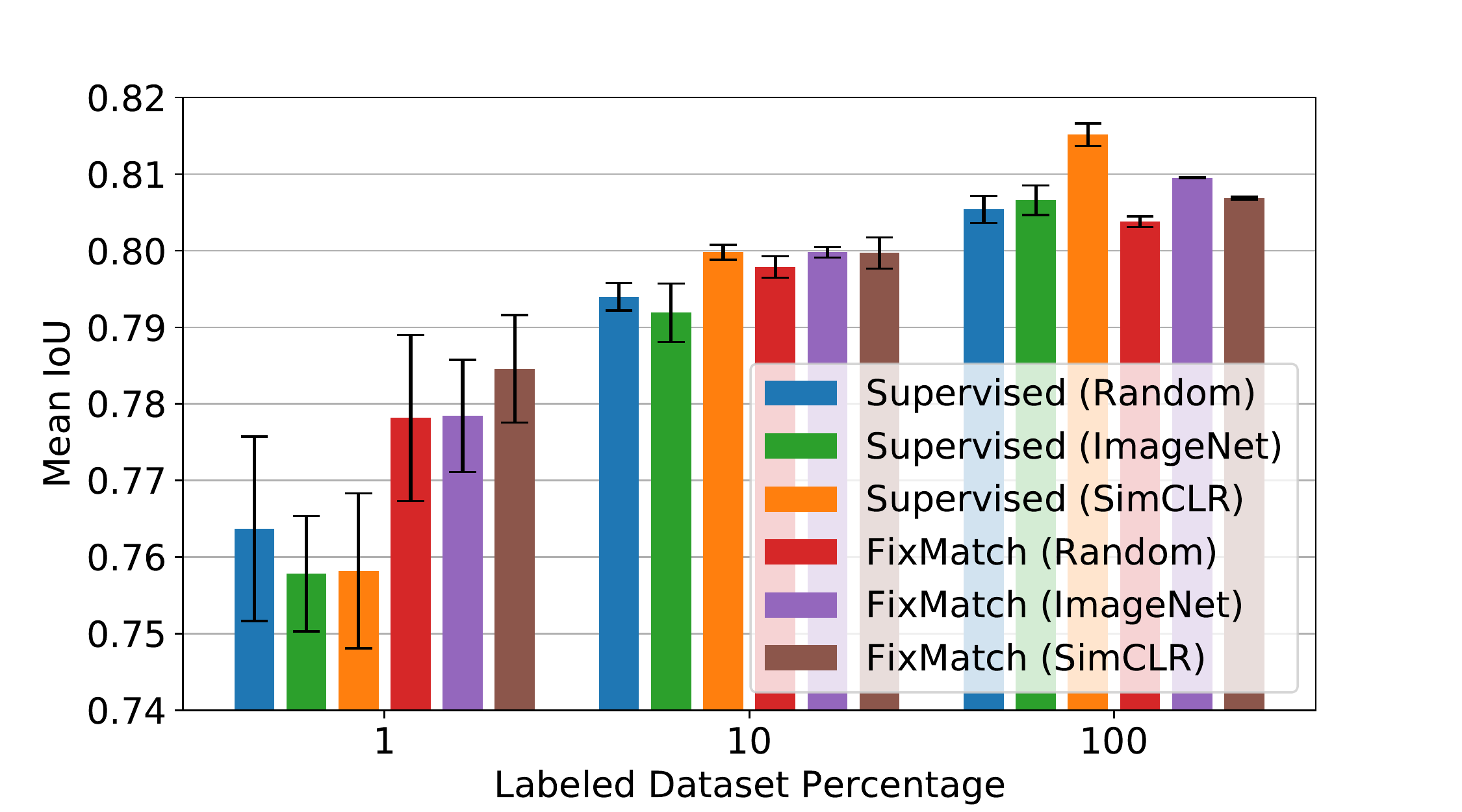}
        \caption{Results on the validation split of Chesapeake domain-shifted partition (left) and IID partition (right).}
        \label{fig:results_chesa_val}
    \end{subfigure}
    \begin{subfigure}
        \centering
        \includegraphics[width=0.45\linewidth]{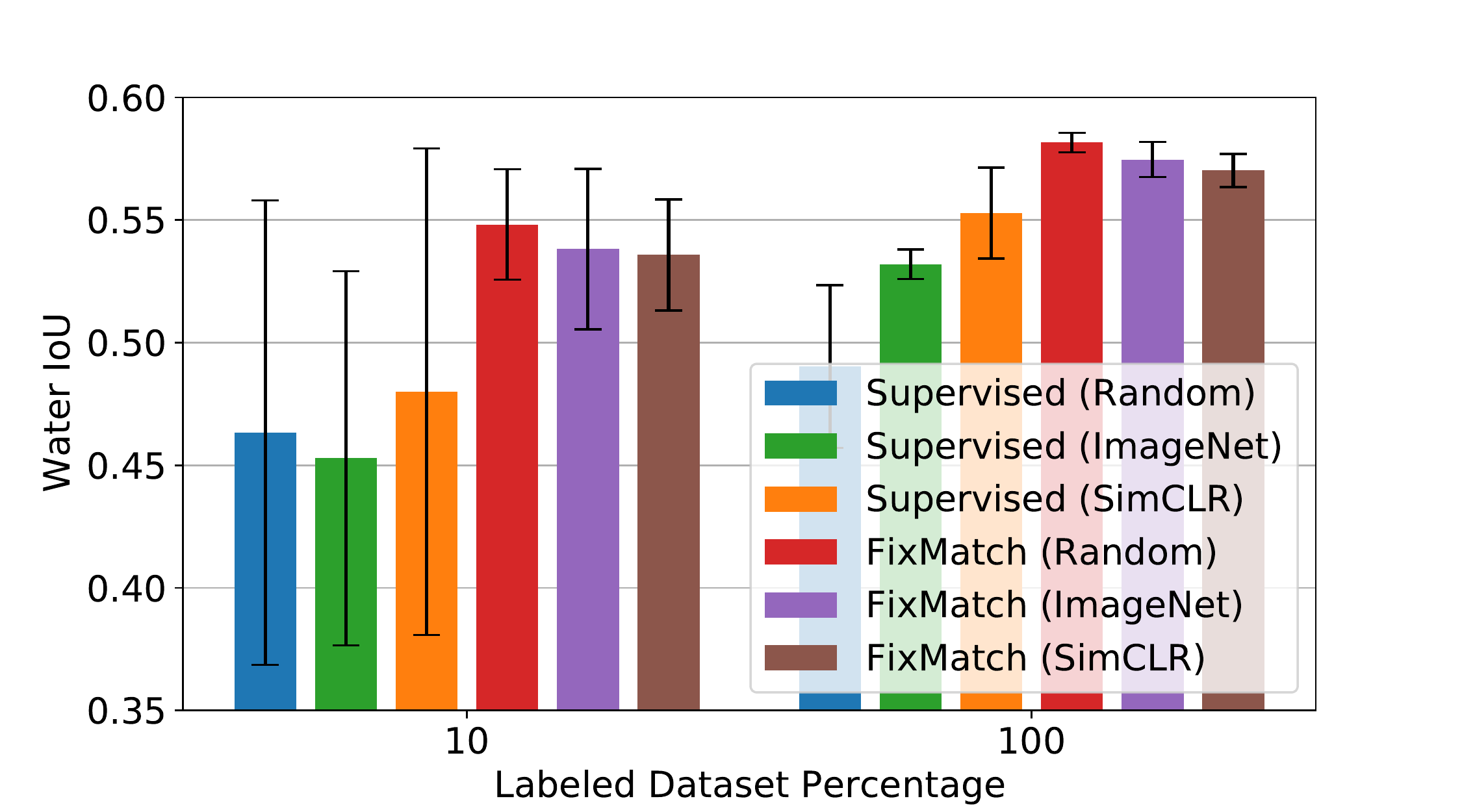}
        \includegraphics[width=0.45\linewidth]{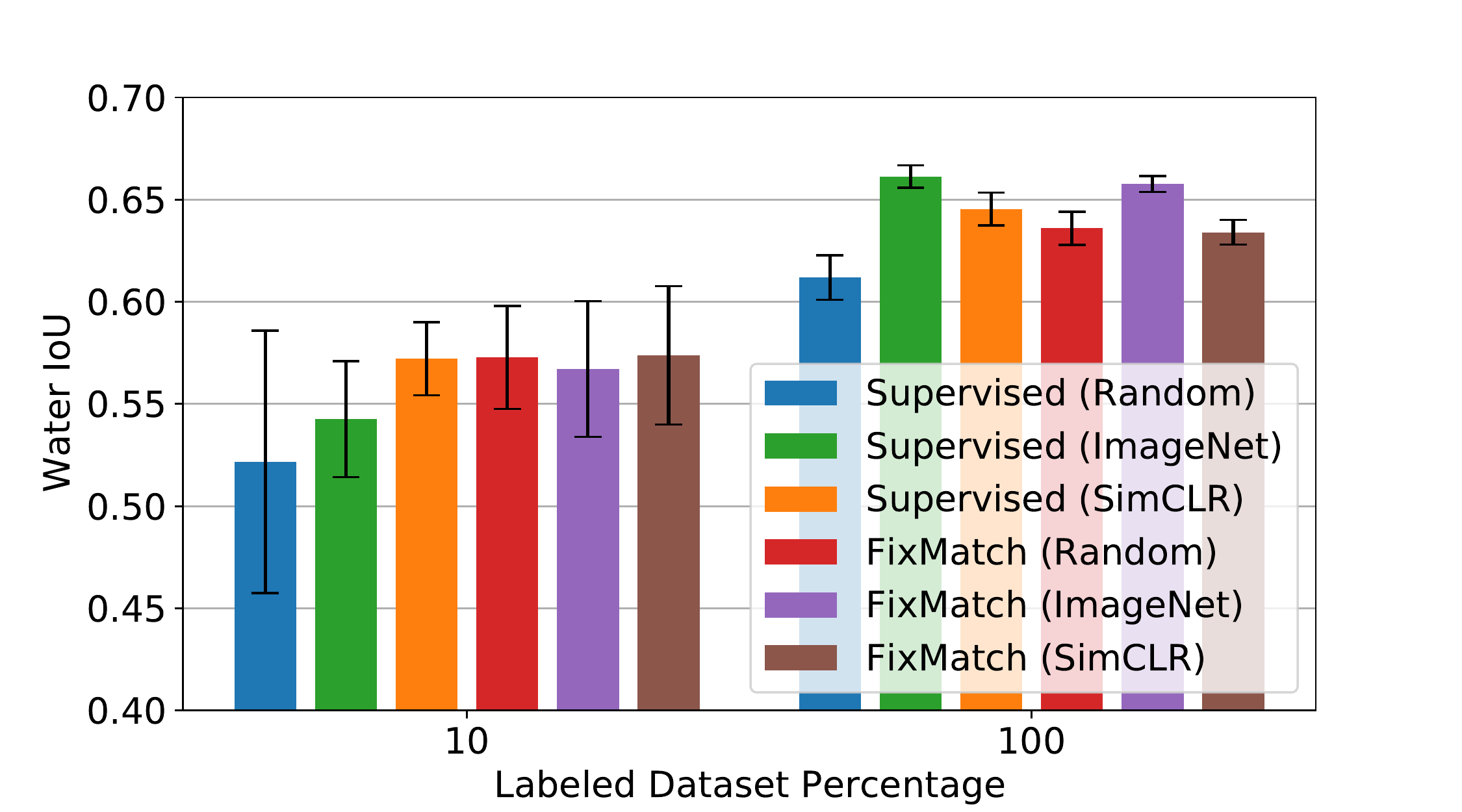}
        \caption{Results on the validation split of Sen1Floods11 domain-shifted partition (left) and IID partition (right).}
        \label{fig:results_sen1floods11_val}
    \end{subfigure}
\end{figure*}

Figures \ref{fig:results_river_val}, \ref{fig:results_chesa_val}, \ref{fig:results_sen1floods11_val} show the results of experiments on the selected hyperparameters on the validation split of the datasets.